\documentclass[conference]{IEEEtran}
\IEEEoverridecommandlockouts
\usepackage{xcolor}
\usepackage{tcolorbox}
\usepackage{alltt}
\usepackage{array, makecell}
\usepackage[linesnumbered,ruled,vlined]{algorithm2e}

\SetCommentSty{mycommfont}

\usepackage{cite}
\usepackage{amsmath,amssymb,amsfonts}
\usepackage{algorithmic}
\usepackage{graphicx}
\usepackage{textcomp}
\usepackage{xcolor}
\def\BibTeX{{\rm B\kern-.05em{\sc i\kern-.025em b}\kern-.08em
    T\kern-.1667em\lower.7ex\hbox{E}\kern-.125emX}}

\usepackage{float}                  
\usepackage{subfig}                 
\usepackage{overpic}                
\usepackage{multirow}
\usepackage{adjustbox}
\usepackage{hyperref}
\usepackage{csquotes}
\usepackage{xcolor}
\usepackage{amsmath}
\DeclareMathOperator*{\argmax}{arg\,max}
\DeclareMathOperator*{\argmin}{arg\,min}
\definecolor{ForestGreen}{RGB}{34,139,34}

\newcommand{\dqi}[1]{{{\textcolor{orange}{\{\{DQi: \bf #1\}\}}}\xspace}}
\newcommand{\sqlgen}{{\sc{FeatAug}}\xspace}

\newtheorem{definition}{Definition}
\newtheorem{problem}{Problem}
\newtheorem{example}{Example}

\begin{document}

\title{FeatAug: Automatic Feature Augmentation From One-to-Many Relationship Tables
}

\author{\IEEEauthorblockN{Danrui Qi}
\IEEEauthorblockA{
\textit{Simon Fraser University}\\
Vancouver, Canada \\
dqi@sfu.ca}
\and
\IEEEauthorblockN{Weiling Zheng}
\IEEEauthorblockA{
\textit{Simon Fraser University}\\
Vancouver, Canada \\
weiling\_zheng@sfu.ca}
\and
\IEEEauthorblockN{Jiannan Wang}
\IEEEauthorblockA{
\textit{Simon Fraser University}\\
Vancouver, Canada \\
jnwang@sfu.ca}
}

\maketitle

\begin{abstract}
Feature augmentation from one-to-many relationship tables is a critical but challenging problem in ML model development. To augment good features, data scientists need to come up with SQL queries manually, which is time-consuming. \textit{Featuretools}~\cite{FeatureTools} is a widely used tool by the data science community to automatically augment the training data by extracting new features from relevant tables. It represents each feature as a group-by aggregation SQL query on relevant tables and can automatically generate these SQL queries. However, it does not include predicates in these queries, which significantly limits its application in many real-world scenarios. To overcome this limitation, we propose \sqlgen, a new feature augmentation framework that automatically extracts predicate-aware SQL queries from one-to-many relationship tables. This extension is not trivial because considering predicates will exponentially increase the number of candidate queries. As a result, the original \textit{Featuretools} framework, which materializes all candidate queries, will not work and needs to be redesigned. We formally define the problem and model it as a hyperparameter optimization problem. We discuss how the Bayesian Optimization can be applied here and propose a novel warm-up strategy to optimize it. To make our algorithm more practical, we also study how to identify promising attribute combinations for predicates. We show that how the beam search idea can partially solve the problem and propose several techniques to further optimize it. Our experiments on four real-world datasets demonstrate that FeatAug extracts more effective features compared to \textit{Featuretools} and other baselines. The code is open-sourced at \url{https://github.com/sfu-db/FeatAug}.
\end{abstract}


\section{Introduction}

Machine learning (ML) can be applied to tackle a variety of important business problems in the industry, such as customer churn prediction~\cite{vafeiadis2015comparison}, next purchase prediction~\cite{liu2016repeat}, and loan repayment prediction~\cite{malhotra2003evaluating}. While promising, the success of an ML project highly depends on the availability of good features~\cite{domingos2012few}. When training data does not contain sufficient signals for a learning algorithm to train an accurate model, there is a strong need to investigate how to augment new features. 

\sloppy

\subsection{Motivation}
Due to the handcrafted feature augmentation being time-consuming, many automatic feature augmentation methods including ~\cite{ExploreKit, AutoCross, autofeat, OpenFE, FETCH, FC-Tree, SAFE, LearningFE, ARDA, FARL, FeatureTools, tsfresh} have been proposed. Most of them focus on extracting augmented features from the training table itself. However, in practice, there is relevant information stored in other tables that can be used to augment the training table.

\fussy

\begin{figure}[t]\vspace{0em}
  \centering
\includegraphics[width=\linewidth]{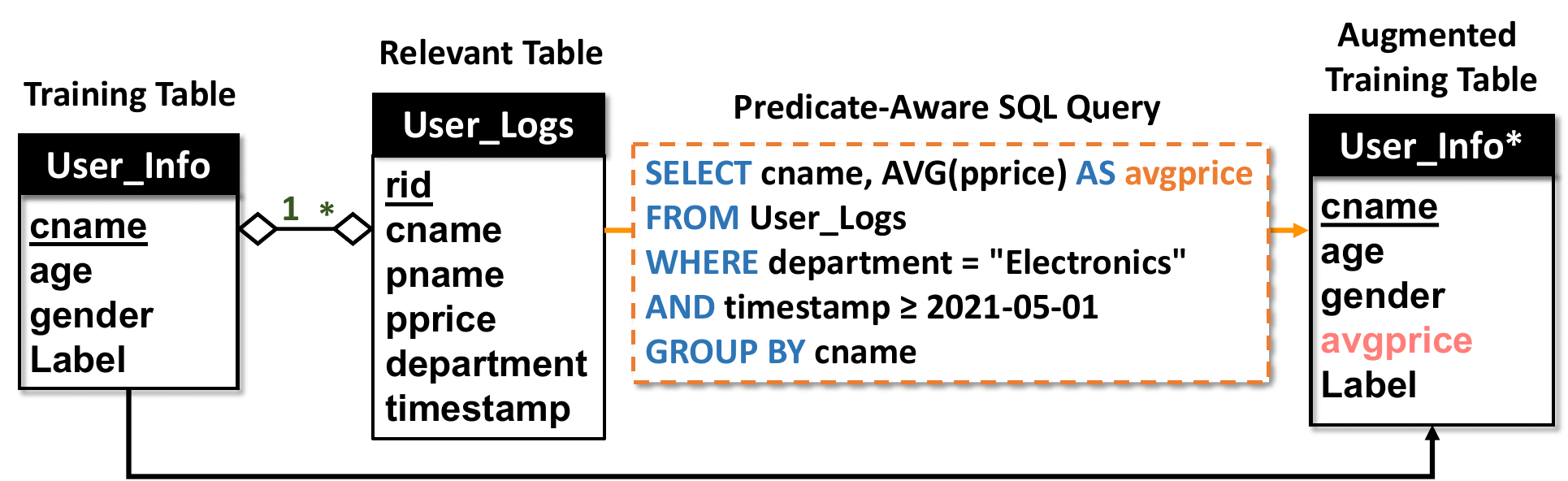}
  \caption{Feature augmentation with predicate-aware SQL queries.}
  \label{fig:running_example} 
\end{figure}

\begin{example}\label{exa:prior_motivation}
Consider a scenario for predicting a customer's next purchase. We want to use customer data from the past 12 months (August 1st, 2022 to July 31st, 2023) to predict whether a customer will purchase a Kindle in August 2023. We have a training table called \texttt{User\_Info} and a relevant table called \texttt{User\_Logs} (shown in Figure~\ref{fig:running_example}). The \texttt{User\_Info} table contains only a limited number of potentially useful features (i.e. age and gender), so additional useful features (i.e. avgprice) should be extracted from the \texttt{User\_Logs} table.
\end{example}

However, we cannot simply add the columns of \texttt{User\_Logs} to \texttt{User\_Info} because the two tables have a one-to-many relationship. That is, each row in \texttt{User\_Info} represents a customer, and a customer may have multiple purchases in \texttt{User\_Logs}. To handle a one-to-many relationship table, data scientists typically write aggregation queries on relevant tables to extract features, which is time-consuming.  \textcolor{blue}{Note that the one-to-many relationship is important. When the training table has one-to-one or many-to-one relationships with a relevant table, a direct join can be used for augmenting features. For many-to-many relationships, they can be divided into many-to-one and one-to-many relationships. The focus then is on addressing the one-to-many relationships through aggregation queries with predicates. 
}




\begin{example}\label{exa:motivation}\it
Continuing with Example~\ref{exa:prior_motivation}, the relationship between \texttt{User\_Info} and \texttt{User\_Logs} is one-to-many with the foreign key \texttt{cname}. To predict whether a customer will purchase a Kindle in August 2023, a data scientist may believe that the amount a customer spent in the past is related to the likelihood that the customer will purchase a Kindle in the future. Consequently, she may write the following aggregation query to generate a feature:

\begin{alltt}
  SELECT cname, AVG(pprice) AS feature
  FROM User\_Logs
  GROUP BY cname
\end{alltt}%

To extract more useful features, the data scientist may need to write multiple queries by considering other aggregation functions such as \textit{COUNT} and other columns in User\_Logs like \texttt{pname}, which can be tedious and time-consuming.
\end{example}

Being able to \emph{automatically} generate features from a one-to-many relationship table will facilitate various ML applications. The following example illustrates how Featuretools [22], widely used in the data science community, addresses this issue.

\begin{example}\label{exa:motivation2}\it 
To relieve data scientists from the tedious task of writing SQL queries, Featuretools generates features automatically by constructing SQL queries in the following format:

\begin{alltt}
  SELECT cname, agg(a) AS feature 
  FROM User\_Logs
  GROUP BY cname
\end{alltt}

Here, \texttt{agg} is an aggregation function such as SUM, AVG, and MAX, and \texttt{a} is an attribute in \texttt{User\_Logs} used for aggregation. By joining the query result with the training table \texttt{User\_Info} on \texttt{cname}, new features can be added to the training data. 

\end{example}

One significant limitation of Featuretools\cite{FeatureTools} is that it does not take predicates into account when generating queries. However, users' interests change rapidly, and purchases made on specific days like "Black Friday" or "Double 11" have little long-lasting impact on sales. Thus, extracting features by aggregating user behavior logs within a specific time slot rather than using all logs is more helpful.


\begin{example}\label{exa:motivation-cond}\it
Continuing Example~\ref{exa:motivation},  Featuretools cannot automatically generate the following SQL query with predicates named predicate-aware SQL query:
\begin{alltt}
  SELECT cname, AVG(pprice) AS avgprice 
  FROM User\_Logs
  WHERE department = "Electronics" 
  AND timestamp \(\geq\) 2023-07-01
  GROUP BY cname
\end{alltt}
In fact, this is a useful feature. The more money a customer spends on ``Electronics'' products in a recent month, the more likely the customer will purchase a Kindle next month. 

\end{example}

\subsection{Chanllenges and Our Methodology} Obviously, it is impossible to materialize all predicate-aware SQL queries because of two reasons: 
\begin{itemize}
    \item \textit{(R1)} The number of SQL queries that can be constructed is huge even though the attribute combination in \texttt{WHERE} clause is fixed.
    \item \textit{(R2)} There is not only one attribute combination that can form the \texttt{WHERE} clause. 
\end{itemize}

\emph{Thus, can we directly find useful SQL queries (i.e. features) from the large search space?} Our key idea is to ``learn'' which areas in the search space are promising (or not promising), and then prune unpromising areas and generate SQL queries from promising areas. \
Based on this idea, we propose \sqlgen, a predicate-aware SQL query generation framework. Given a training table and relevant table, \sqlgen aims to automatically extract useful features from the relevant table by constructing predicate-aware SQL queries. \sqlgen contains two components to filter out unpromising queries. 

 \textit{\textbf{For R1}}, \sqlgen needs to search for promising predicate-aware SQL queries by searching for the proper aggregation function, attributes for aggregation and values that can fill out the 
\texttt{WHERE} clause. Our key idea is to model the SQL query generation problem as a hyperparameter optimization problem. 
By modelling the correlation between the SQL queries (i.e. features) and their performance, we introduce an exploration-and-exploitation strategy to enhance the search process. Moreover, we also warm up the search process by transferring the knowledge of related tasks.

\textit{\textbf{For R2}}, \sqlgen need to search for promising attribute combinations in \texttt{WHERE} clause. Our key idea is to model the search space as a tree-like search space and greedily expand the tree by predicting the performance of each tree node. Note that each tree node represents an attribute combination. To reduce the long evaluation time of each tree node, we take the low-cost proxy to simulate the real evaluation score.

\subsection{Our Contributions} We make the following contributions in this paper:

\begin{itemize}
    \item We study a novel predicate-aware SQL query generation problem of automatic feature augmentation from one-to-many relationship tables motivated by real-world ML applications. We formally define the \textit{Predicate-Aware SQL Query Generation} problem. 

    \item We develop \sqlgen, a predicate-aware SQL query generation framework to enable automatic feature augmentation from one-to-many relationship tables.

    \item  We model the problem of searching for promising predicate-aware SQL queries as a hyperparameter optimization problem. We enhance the search process by introducing exploration-and-exploitation strategy and transferring the knowledge of related tasks.

    \item  We model the search space of promising attribute combinations in \texttt{WHERE} clause as a tree-like space and greedily expand it by predicting the performance of each attribute combination.

    \item The empirical results on four real-world datasets show the effectiveness of \sqlgen on both traditional and deep ML models.
    Compared to the popular \textit{Featuretools} and other baselines with the same number of generated features, \sqlgen can get up to 10.74\% AUC improvement on classification tasks and 0.0740 RMSE improvement on regression tasks.
\end{itemize}

\section{Related Work}\label{sec:rw}
In this section, we present the related work of this paper from four perspectives including automatic feature augmentation, feature selection in automatic feature augmentation, data enrichment and hyperparameter optimization.

\subsection{Automatic Feature Augmentation.} There are several existing efforts on automated feature augmentation~\cite{ExploreKit, AutoCross, autofeat, OpenFE, FETCH, FC-Tree, SAFE, LearningFE, ARDA, FARL, FeatureTools, tsfresh, AutoFP}. However, they complement our work and focus on different scenarios. Explorekit~\cite{ExploreKit}, FC-Tree~\cite{FC-Tree}, SAFE~\cite{SAFE}, LFE~\cite{LearningFE}, Auto-Cross~\cite{AutoCross}, Autofeat~\cite{autofeat}, OpenFE~\cite{OpenFE}, and FETCH~\cite{FETCH} work for the single table scenario. \textcolor{blue}{They automatically generate new features by applying unray operators, binary operators and feature crossing operations to existing features,} which is orthogonal to the scenario \sqlgen applies. ARDA~\cite{ARDA} and AutoFeature~\cite{FARL} fit the scenarios with multiple relational tables by assuming that each table can be directly joined with the training table, i.e. the one-to-one relationship tables. 
\textcolor{blue}{ARDA automatically joins the top relevant tables with the base table according to the relevant score it computed, While AutoFeature aims to filter out the effective feature set from the relevant tables that can be joined with the base table. }
Our work mainly considers the one-to-many relationship tables, which cannot be solved by directly joining. Featuretools~\cite{FeatureTools} is a popular tool that automatically augments new features for one-to-many relationship tables. \textcolor{blue}{It augments new features to the training table through generating SQL queries by using aggregation functions such as \textsf{SUM, COUNT} and \textsf{MIN} without considering predicates.} In contrast, \sqlgen is predicate-aware, i.e. \sqlgen considers predicates in the \texttt{WHERE} clause when generating SQL queries. 

\subsection{Feature Selection in Automatic Feature Augmentation.} Feature selection~\cite{chandrashekar2014survey, guyon2003introduction} aims to only keep effective features and filter out features that have little or even negative impact on the performance of the downstream ML model. Several automatic feature augmentation methods follow the expand-and-reduce framework and utilize different feature selection strategies. \textcolor{blue}{FC-Tree~\cite{FC-Tree} and SAFE~\cite{SAFE} use the information gain such as mutual information to select useful features. AutoCross~\cite{AutoCross} and AutoFeat~\cite{autofeat} use the improvement of a linear regression model to evaluate whether a feature is effective. } LFE~\cite{LearningFE} and ExploreKit~\cite{ExploreKit} use meta-features to train an ML model and predict whether a new coming feature is effective. \textcolor{blue}{AutoFeature~\cite{FARL} defines a reward function, i.e. the improvement of an XGBoost model to measure the impact of adding a feature on the performance of a downstream ML model.} Different from the previous methods, \sqlgen filters out useless features by identifying promising query templates and promising search areas in query pools prior.

\subsection{Data Enrichment.} Data enrichment aims to augment a local table with new attributes extracted from external data, such as Web Tables~\cite{yakout2012infogather,cafarella2009data,fan2014hybrid} and Deep Web~\cite{wang2019progressive, zhao2020activedeeper, wang2018deeper}. \textcolor{blue}{In Deep Web, the knowledge from deep web (i.e., a hidden database) are progressively crawled through a keyword-search API
to enrich a local table. These data enrichment methods do not measure the practical impact of enriched data on the particular ML model i.e. they are not model-aware.} While \sqlgen is a feature augmentation method that is model-aware, i.e. the enrichment goal of \sqlgen is to maximize the performance of the downstream ML model.

\subsection{Hyperparameter Optimization.} Hyperparameter Optimization has been extensively studied in the ML community~\cite{hutter2019automated}. Random Search~\cite{Random-Search} is one simple but effective method in this area. \textcolor{blue}{Bayesian Optimization (BO) is another popular methodology in this area using one surrogate model to establish the correlation between hyperparameters and the downstream ML model performance. Establishing these surrogate models is often expensive. The commonly used surrogate models include Gaussian Process (GP)~\cite{SMBO, schonlau_global_1998}, Tree-structured Parzen Estimators (TPE)~\cite{pmlr-v28-bergstra13, NIPS2011_86e8f7ab}, and Random Forest~\cite{hutter2011sequential}. To speed up the search process of BO, Hyperband~\cite{Hyperband} and BOHB~\cite{BOHB} are proposed with the parallel downstream ML model fitting and the early-stopping ideas. } Our work proposes a novel framework to bridge hyperparameter optimization and predicate-aware SQL query, i.e. feature generation, and may open up a new avenue for future research in automatic feature augmentation.

\begin{figure}[t]\vspace{-2em}
  \centering  \includegraphics[width=\linewidth]{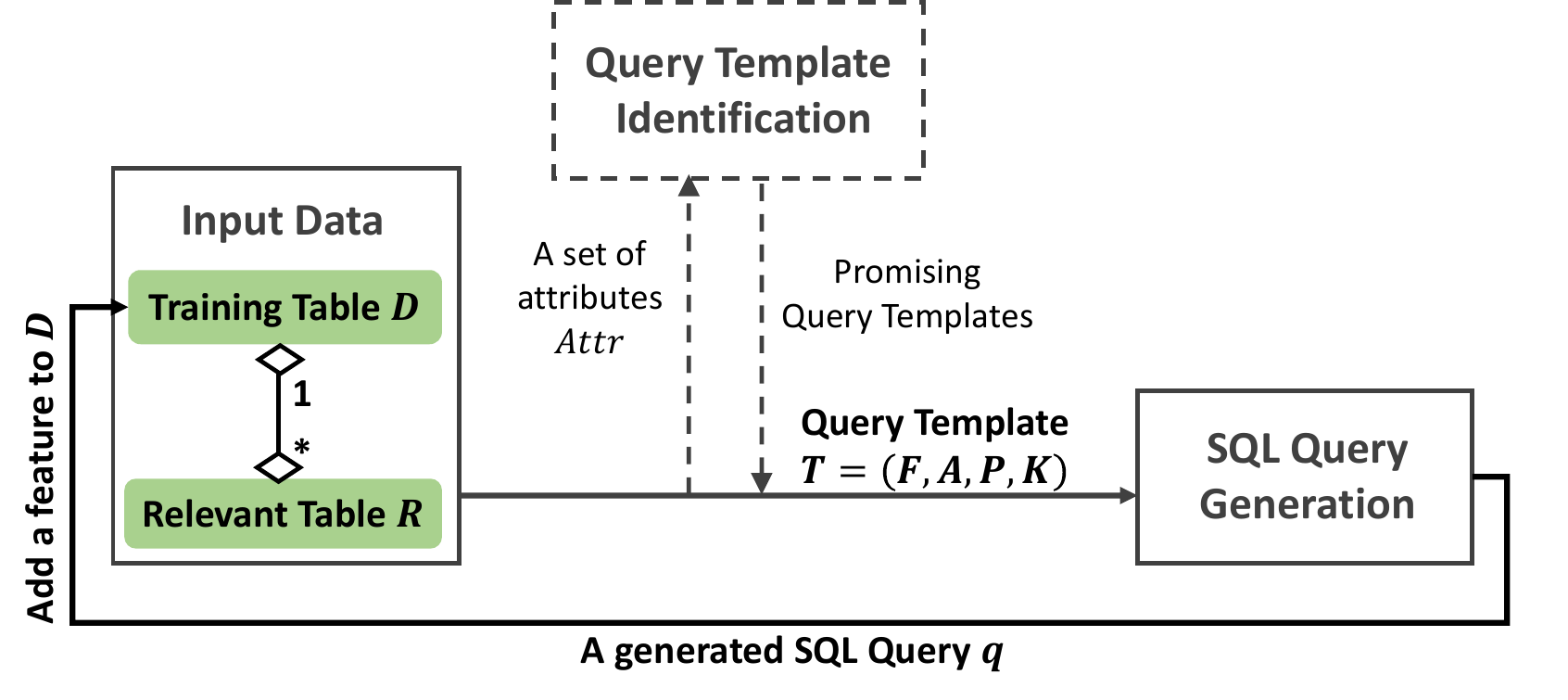}
  \caption{Workflow of \sqlgen.}
  \label{fig:SQLGENFramework} \vspace{-1em}
\end{figure}

\section{Problem Formulation}
In this section, we formulate two problems including \textit{Predicate-Aware SQL Query Generation} and \textit{Query Template Identification}.

\subsection{Predicate-Aware SQL Query Generation}
We formulate the Predicate-Aware SQL query generation problem for one-to-many relationship tables. Without losing generalization, we first define the problem under the scenario with one base table and one relevant table. It is easy to be extended to more complex scenarios. For the \textit{Deep-Layer Relationships}~\cite{FeatureTools}, it can be represented by the aforementioned scenario by joining all the tables into one relevant table. For the scenario with multiple relevant tables, it can be represented by multiple scenarios with one base table and one relevant table.



Let $D$ denote a \emph{training table}, which has a primary key, a set of features, and a label. Let $R$ denote a \emph{relevant table} contains the foreign key referring to $D$'s primary key. Firstly, we define \emph{Query Template} used to generate effective predicate-aware SQL queries:

\begin{definition}[Query Template] \label{def:query-template} Given a relevant table R with a set of attributes $Attr = \{A_1, A_2, \cdots, A_m\}$ where $A_i$ denotes the $i$-th attribute of $R$, a query template w.r.t. $R$ is a quadruple $T = (F, A, P, K)$, where $F$ is a set of aggregation functions, $A \subseteq Attr$ is a set of attributes which can be aggregated, $P \subseteq Attr$ is a fixed attribute combination forming \texttt{WHERE} clause, and $K$ is the foreign key attributes.
\end{definition}

\begin{example}\label{exa:q-temp}
Consider the relevant table in Example~\ref{exa:prior_motivation}. Here is an example query template w.r.t. the table: 

{\footnotesize
\begin{align*}
T = \Big([\texttt{SUM, AVG, MAX}], [\texttt{pprice}], [\texttt{department}, \texttt{timestamp}],
[\texttt{cname}]\Big)
\end{align*}
}

\noindent where [\texttt{SUM, AVG, MAX}] is the aggregation function set $F$, $A = 
 $[\texttt{pprice}] is the set of attributes for aggregation, [\texttt{depar-}
 \texttt{tment}, \texttt{timestamp}] is the fixed attribute combination that forms the \texttt{WHERE} clause, and [\texttt{cname}] is the foreign key attribute between $D$ and $R$.
\end{example}


A query template represents a pool of candidate SQL queries. Definition~\ref{def:query-pool} defines the query pool w.r.t. a given query template.

\begin{definition}[Query Pool]\label{def:query-pool}
Given a relevant table $R$ and a query template $T = (F, A, P, K)$, a query pool $Q_T$ consists of a collection of predicate-aware SQL queries in the following form: 
\begin{alltt}
  SELECT k, agg(\(a\)) AS feature FROM \(R\)
  WHERE predict(\(p\sb{1}\)) AND ... AND predict(\(p\sb{w}\))
  GROUP BY k
\end{alltt}
where agg $\in F$, $a \in A$, $p_i\in P$ for each $i\in[1, w]$ and $k \subseteq K$ is a subset of the foreign key attributes. If $p_i$ is a categorical column, predicate$(p_i)$ represents an equality predicate, i.e. $p_i = d$, where $d$ is a value in the domain of $p_i$; if $p_i$ is a numerical or datetime column, predicate$(p_i)$ represents a range predicate, $d_{low} \leq p_i \leq d_{high}$, where $d_{low}$ and $d_{high}$ are two values in the domain of $p_i$ ($d_{low} \leq d_{high}$). Note that the range-predicate definition includes one-sided range predicates. 
\end{definition}

\begin{example}
Continuing Example~\ref{exa:q-temp}, the query pool $Q_T$ related to the query template $T$ is composed of predicate-aware SQL queries in the following form. And the SQL query in Example~\ref{exa:motivation-cond} is one query in $Q_T$.
\begin{alltt}
  SELECT cname, agg(\(a\)) AS feature 
  FROM User\_Logs 
  WHERE department = '?' 
  AND timestamp \(\geq\) '?' AND timestamp \(\leq\) '?'
  GROUP BY cname
\end{alltt}

\end{example}


For a predicate-aware SQL query $q \in Q_T$, let $q(R)$ denote the result table by executing $q$ on $R$. Definition~\ref{def:feat-aug} defines the augmented training table that adds the generated feature in $q(R)$.

\begin{definition}[Augmented Training Table]\label{def:feat-aug}
Given a training table $D$ and a query result table $q(R)$, the augmented training table $D^q$ is defined as:
\begin{alltt}
  SELECT \(D\).*, \(q(R)\).feature 
  FROM \(D\) LEFT JOIN \(q(R)\) 
  ON D.k = \(q(R)\).k
\end{alltt}
\end{definition}

\begin{example}
Continuing Example~\ref{exa:motivation-cond}, after executing the query in Example~\ref{exa:motivation-cond}, we get the query result table $q$(\texttt{User\_Logs})=
(\texttt{\underline{cname}, avgprice}). We can get the augmented training table by joining \texttt{User\_Info} with $q$(\texttt{User\_Info}) and get $D^q$=(\texttt{\underline{cname}, age, gender, avgprice, label}) with the following SQL query:
\begin{alltt}
  SELECT User\_Info.*, q(User\_Logs).avgprice 
  FROM User_Info 
  LEFT JOIN q(User\_Logs) 
  ON User\_Info.cname = q(User\_Logs).cname
\end{alltt}
\end{example}

To evaluate the effectiveness of generated SQL query, i.e. feature, $D^{q}$ can be split into a training set $D^{q}_{train}$ and a validation set $D^{q}_{valid}$, where $D^{q}_{train}$ is used to train an ML model and $D^{q}_{valid}$ is used to evaluate model performance. The lower the model loss, the more effective the generated SQL query, i.e. feature. 

The goal of \textit{Predicate-Aware SQL Query Generation Problem} is to minimize the model loss by generating effective predicate-aware SQL queries, i.e. features. This implies an optimization problem. We formally define the \textit{Predicate-Aware SQL Query Generation Problem}.

\begin{problem}[Predicate-Aware SQL Query Generation]\label{def:feat-dis}
\textit{Given a training table $D$, a relevant table $R$, a query template $T$ and an ML model $\mathcal{A}$, the goal of predicate-aware query generation is to find the most effective query $q^* \in Q_T$ such that the model trained on $D^{q^*}_{train}$ and evaluated on $D^{q^*}_{valid}$ achieves the lowest loss, i.e., }
\begin{align*}
 q^*= \argmin_{q \in Q_T} \mathcal{L}(\mathcal{A}(D^{q}_{train}), D^{q}_{valid}),
\end{align*}
where $\mathcal{A}(D^{q}_{train})$ represents the model trained on the training set $D^{q}_{train}$. $\mathcal{L}(\cdot, \cdot)$ takes a model and the validation set $D^{q}_{valid}$ as input, and returns the validation loss of the trained model. 
\end{problem}

\subsection{Query Template Identification}
In practice, users who are unfamiliar with the data often cannot provide explicit query templates for generating effective SQL queries. To deal with the more general scenario, we need to identify the query templates that are useful for generating effective SQL queries. Thus, we formally define the \textit{Query Template Identification} problem.

\begin{definition}[Query Template Set] \label{def:query-template-set}
Given a relevant table R and an attribute set $Attr = \{A_1, A_2, \cdots, A_m\}$ where $A_i$ denotes the $i$-th attribute of 
R, a query template set $\mathcal{S}$ w.r.t. R is a set including all possible query templates, i.e. $\mathcal{S} = \{(F, A, P, K) | \forall P \subseteq Attr\}$.
\end{definition}

\begin{example}\label{exa:query-template-set}
Consider the query template $T$ in Example~\ref{exa:q-temp}. There are other query templates by differentiating $P$, i.e. [\texttt{dep-}
\texttt{artment}, \texttt{timestamp}] in $T$. Here are other two example query templates:

{\footnotesize
\begin{align*}
&T_{1} = \Big([\texttt{SUM, AVG, MAX}], [\texttt{pprice}], [\texttt{pname}, \texttt{pprice}],
[\texttt{cname}]\Big) \\
&T_{2} = \Big([\texttt{SUM, AVG, MAX}], [\texttt{pprice}], [\texttt{pname}, \texttt{department}], [\texttt{cname}]\Big)
\end{align*}
}

There are $2^6$ different query templates, which can be constructed as the query template set $\mathcal{S}$.

\end{example}

\begin{definition}[Effectiveness of Query Template]\label{def:query-template-idf}
Given a training table $D$, a relevant table $R$, a query template $T$ and an ML model $\mathcal{A}$, the effectiveness of query template $T$ is defined as,
\begin{align*}
e_{T} = \mathcal{L}(\mathcal{A}(D^{q^*}_{train}), 
D^{q^*}_{valid})
\end{align*}
where $q^*$ is the most effective SQL query in $Q_T$.
\end{definition}

\begin{problem}[Query Template Identification]\label{def:query-template-idf}
Given a training table $D$, a relevant table $R$ and a set of attributes $attr \subseteq Attr$, the query template set w.r.t. attr is $\mathcal{S}_{attr}=\{(F, A, P, K)|\forall P \subseteq attr\}$. The goal of query template identification is to recommend $n$ query templates $T_1, T_2, \cdots, T_n \in \mathcal{S}_{attr}$, where $T_1, T_2, \cdots, T_n$ shows top-n effectiveness over all query templates in $\mathcal{S}_{attr}$.
\end{problem}

\section{The FEATAUG Framework}

In this section, we present the overview of \sqlgen framework. Then we exhibit more details of each part in \sqlgen.
We first introduce the overview of \sqlgen workflow. Without losing generalization, we exhibit more details of each part by taking the scenario with one base table and one relevant table as an example.  Figure~\ref{fig:SQLGENFramework} illustrates the \sqlgen framework. It takes a training table and a relevant table that has one-to-many relationship with the training table as input. In the \textit{SQL Query Geneartion} component, \sqlgen iteratively searches for the effective queries (i.e. features) in the query pool. Then in the \textit{Query Template Identification Component}, \sqlgen iteratively searches $n$ promising query templates which seem to include effective predicate-aware SQL queries in their query pools.

\subsubsection{Workflow of FeatAug} The \sqlgen workflow is shown in Figure~\ref{fig:SQLGENFramework}. It includes two components named \textit{SQL Query Generation} and \textit{Query Template Identification}. The \textit{SQL Query Generation} component aims to generate effective predicate-aware SQL queries. It takes a training table $D$, a relevant table $R$ and a query template $T$ as input. It outputs an effective predicate-aware SQL query $q$ that can augment one feature into $D$. If users want to get multiple effective SQL queries in $Q_T$, they just need to call the \textit{SQL Query Generation} component multiple times. However, in practice, users who are unfamiliar with the input data often cannot specify the attribute combination $P$ in $T$ explicitly. Instead, they can only provide a set of attributes in $R$ which may construct promising query templates or even nothing. 
Then the \textit{Query Template Identification} component is optional in \sqlgen. It deals with the situation that the users cannot provide explicit query templates. Given one training table $D$, one relevant table $R$ and a set of attributes in $R$, the \textit{Query Template Identification} component aims to figure out $n$ most promising query templates, i.e. $n$ most promising attribute combinations for constructing the \texttt{WHERE} clause. It outputs a set of promising query templates as the input of the \textit{SQL Query Generation} component to generate effective SQL queries.

\subsubsection{The SQL Query Generation Component} The SQL Query Generation component takes a training table $D$, a relevant table $R$ and a query template $T$ as input, and searches for the effective predicate-aware SQL query $q \in Q_T$.
According to the definition of query pool (Definition 3), the \textit{Predicate-Awarer SQL Query Generation} problem can be modelled as the \textit{Hyperparameter Optimization} problem. Thus the query pool $Q_{T}$ can be searched iteratively for generating $q$. 
In this work, \sqlgen utilizes \textit{Bayesian Optimization} as the search strategy, which is commonly employed in the HPO area. At the start of the search process, \sqlgen warms up the search process by transferring knowledge from relative tasks. 

\subsubsection{The Query Template Identification Component} The Query Template Identification Component takes a training table $D$, a relevant table $R$, and a set of attributes $attr$ in $R$ as input. It constructs the query template set $\mathcal{S}_{attr}$ and iteratively searches $n$ promising query templates which seem to include effective predicate-aware SQL queries in their query pools. 
At each iteration, \sqlgen first draws the most promising sample of the query templates from $\mathcal{S}_{attr}$. Then \sqlgen evaluates predicate-aware SQL queries in the related query pools. As Definition 4 shows, the effectiveness of query templates is determined by the
evaluation result of the most effective SQL query in their query pool. Note that the strategy of drawing the most promising query templates is determined by the search strategy \sqlgen employed. In this work, \sqlgen employs the beam-search idea for identifying the most promising query templates layer-by-layer. However, directly applying the beam-search idea is infeasible. We analyze the reason and make it practical in Section VI.

\section{SQL Query Generation}
 In this section, we introduce the \textit{SQL Query Generation} component in \sqlgen.
We first model the \textit{Predicate-Aware SQL Query Generation} problem as the HPO problem. Then, we introduce a representative \textit{Bayesian Optimization} algorithm for executing the search process in the query pool, i.e. \textit{TPE}~\cite{NIPS2011_86e8f7ab}. Finally, we introduce the strategy of warming up the search process.
\subsection{SQL Query Generation as HPO}

In Section 2, we define the \textit{Query Pool} $Q_T$ related to each \textit{Query Template} $T$. Obviously, $Q_T$ is our search space. We first map SQL queries in $Q_T$ into a vector space $\mathcal{V}$, then we can model the \textit{SQL Generation Problem} as \textit{Hyperparameter Optimization Problem}.

Given a query template $T=(F, A, P, K)$ and a SQL query $q \in Q_T$, the corresponding \textit{query vector} $v_q$ consists of four parts: (1) a single element that represents the aggregation function selected from $F$. (2) a single element that represents the attribute for aggregation selected from $A$. (3) a set of possible values for attributes in $P$ forming \texttt{WHERE} clause. Suppose that $P$ contains $n$ categorical attributes and $m$ numerical/datetime attributes, the third part contains $(n+2*m)$ elements because we need 2 elements to represent the range predicate of a numerical/datetime attribute. If the query does not contain a predicate on some attribute, the corresponding element will be set to \textit{None}. Otherwise, the actual value will be shown in the vector. (4) a set of possible values indicating the set of attributes $k$ selected for \texttt{GROUP BY} clause. Note that $k$ is the subset of the foreign key. If one attribute in the foreign key is selected, the corresponding element equals 1, otherwise 0. The fourth part contains $|K|$ elements. By indicating each element of $v_q$ concretely, a query $q \in Q_T$ can be generated. All the \textit{query vector} $v_q \in \mathcal{V}$, i.e. SQL query $q \in Q_T$, constructs the whole search space.

\begin{example}\label{exa:map_SQG_to_HPO}\it
Considering Example~\ref{exa:motivation-cond},
the query vector $v$ corresponding to the SQL query in Example~\ref{exa:motivation-cond} is shown as follows:
\begin{alltt}
    v = [1, 0, 4, '2023-05-01', None, 0]
\end{alltt}

In the query vector, the first element is set to 1, representing the \texttt{AVG} function whose index equals 1 is selected from the aggregation function set [SUM, AVG, MAX]. The second element is set to 0, representing the \texttt{pprice} attribute whose index equals 0 is selected from the aggregation attribute set \texttt{[pprice]}.
The elements from the third place to the fifth place correspond to the attribute combination \texttt{[department, timestamp]} forming the \texttt{WHERE} clause. The third element indicates that the value of \texttt{department} is ``4'', i.e. the encoding of ``Electronics''. Since \texttt{timestamp} is a datetime attribute (occupying two elements in the vector), the fourth element represents the lower bound of \texttt{timestamp} and the fifth element represents the upper bound. The last element denotes the \texttt{cname} attribute whose index equals 0 is selected from the foreign key set \texttt{[cname]}. 
\end{example}

Mapping $Q_T$ to $\mathcal{V}$ provides a natural analogy between the \textit{SQL Generation Problem} and \textit{Hyperparameter Optimization Problem}. In the HPO problem, a set of hyperparameters is also abstracted as a vector $[p_1, p_2, \cdots, p_i]$, and the value of $p_1, p_2, \cdots, p_i$ is picked from the domains of hyperparameters. The goal of the HPO problem is to search for the best vector that achieves the optimal metric. Meanwhile, the \textit{Predicate-Aware SQL Query Generation Problem} also aims to find the most effective SQL queries $v_q \in \mathcal{V}$, i.e. features which lead to minimal validation loss of the ML model.

\begin{example}
Continuing Example~\ref{exa:map_SQG_to_HPO}. The vector in Example~\ref{exa:motivation-cond} is abstracted as a six-dimension vector $v$. 
In the query vector, the domain of the first element is the aggregation function set \texttt{[SUM, AVG, MAX]}. The domain of the second element is the aggregation attribute set \texttt{[pprice]}.
The elements from the third place to the fifth place correspond to the attribute combination \texttt{[department, timestamp]} forming the \texttt{WHERE} clause. Thus, the domain of the third element is the encoding of values in the domain of the \texttt{department} attribute. Since \texttt{timestamp} is a datetime attribute (occupying two elements in the vector), the fourth element represents the lower bound of \texttt{timestamp} and the fifth element represents the upper bound. Thus, the domain of the fourth and fifth element is the domain of the \texttt{timestamp} attribute adding the \texttt{None} value. The domain of the last element is the foreign key set \texttt{[cname]}. The goal of the Predicate-Aware SQL Query Generation Problem is to pick up effective values in the domains of the query vector, which naturally analogies to the HPO Problem. 
\end{example}

\subsection{BO for SQL Query Generation}
In the realm of \textit{Bayesian Optimization (BO)}, the objective is to identify an optimal point $x*$ within a search space $X$, which maximizes an objective function $f$: 
\begin{align*}
x^{*} = \argmax_{x \in \mathcal{X}} f(x).
\end{align*}

Here, $f$ serves as a black-box function lacking a straightforward closed-form solution. This framework is particularly popular in HPO, where $x$ represents a set of hyperparameters and $f(x)$ quantifies the performance of a model governed by those hyperparameters.

BO framework~\cite{SMBO} treats $f$ as an oracle and iteratively queries it to refine a Gaussian Process (GP) surrogate model. Due to the high computational cost of oracle evaluations, acquisition functions like Expected Improvement (EI) are employed to judiciously select subsequent query points. However, the GP surrogate model struggles with discrete points due to GP's inherent properties.

\textit{Tree-structured Parzen Estimator (TPE)} addresses these limitations by utilizing Kernel Density Estimation (KDE) as its surrogate model. It partitions points into "good" and "bad" groups. The boundray between good points and bad points is denoted by $\gamma$ which is some quantile of observed evaluations. $\gamma$ typically equals to 10\%-15\%, i.e. 10\%-15\% points are good points. Then it calculates EI as a function of the ratio 
$P_{good}(x)/P_{bad}(x)$. In multi-dimensional scenarios, a specific KDE model is constructed for each dimension. There are certainly some other optimization approaches~\cite{schonlau_global_1998,pmlr-v28-bergstra13, NIPS2011_86e8f7ab,hutter2011sequential, li2017hyperband}. We adopt TPE in our study for three principal reasons: (1) it has an established reputation in the field of HPO~\cite{NIPS2011_86e8f7ab,pmlr-v28-bergstra13} (2) 
it is more efficient compared to other BO approaches such as SMBO~\cite{SMBO} and SMAC~\cite{SMAC} (3)
TPE is good at optimizing both discrete and continuous hyperparameters.

\textcolor{blue}{\textbf{Remark.} The focus of this work is to demonstrate the applicability of HPO methods to the generation of predicate-aware SQL queries. We choose the popular TPE to achieve this goal. There exists other HPO methods such as SMAC~\cite{SMAC} and BOHB~\cite{BOHB}. It will be interesting to investigate which HPO method is better in the future study. }
\subsection{Warm-up the Surrogate Model }
\subsubsection{Potential Issues of TPE} At the initial stage of \textit{TPE}, it randomly draws a sample of SQL queries from the search space to identify a promising area and exploit this area by selecting the queries around this area. Our problem has a large search space and an expensive evaluation, thus directly applying \textit{TPE} requires a large number of iterations to identify promising areas, which could be very expensive.  This issue will be further exaggerated when the size of training data is large. 



\subsubsection{Our Solution - Warm-up the Surrogate Model} 
\begin{figure}[t]\vspace{0em}
  \centering
  \includegraphics[width=\linewidth]{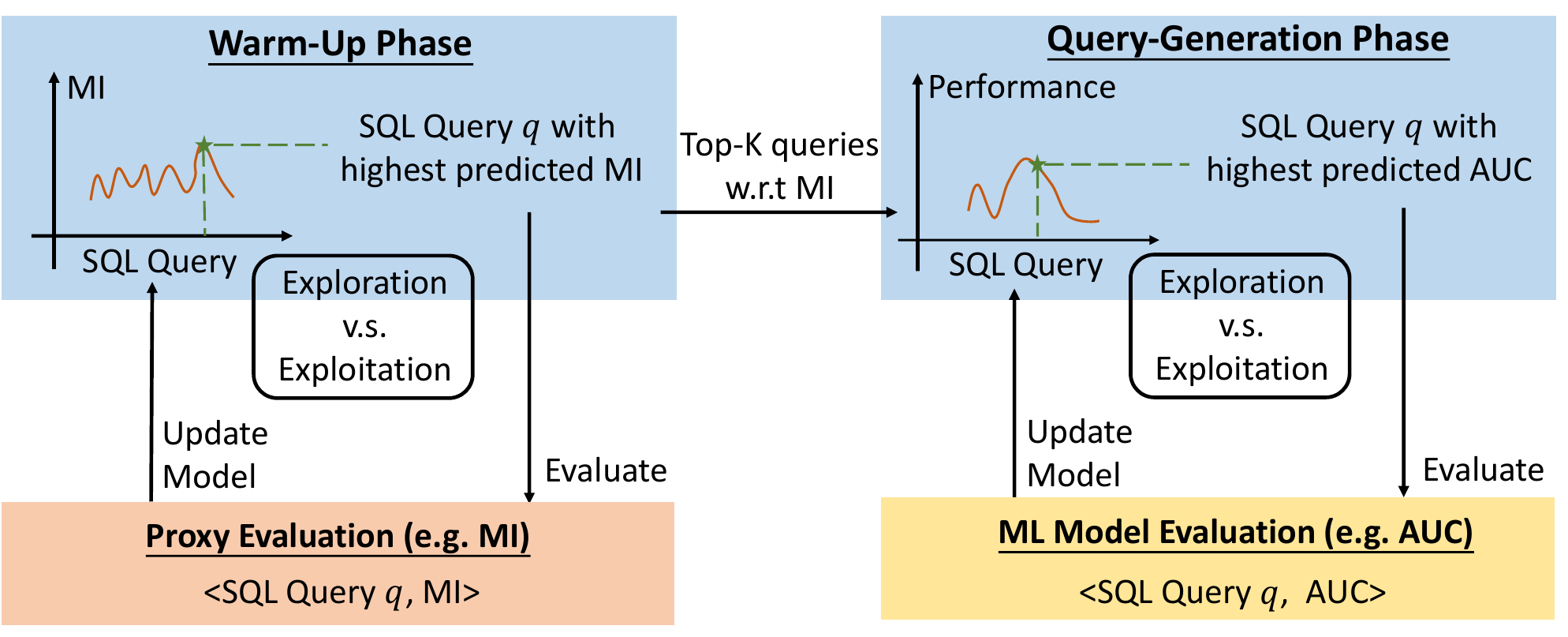}
  \caption{Workflow of SQL Query Generation Component. Mutual Information (MI) is taken as the low-cost proxy.}
  \label{fig:sql-query-generation} 
\end{figure}

Instead of randomly initializing the search process of TPE, we consider transferring the knowledge of the related low-cost tasks to strengthen the initialization, i.e. construct better \textit{KDE}s at the start of \textit{TPE} search. The knowledge-transferring idea can speed up the search process for predicate-aware SQL queries or even get better SQL queries. As shown in Figure~\ref{fig:sql-query-generation}, to incorporate the knowledge of related low-cost tasks into the search process, we propose to run TPE for two rounds. In the first round (Warm-Up Phase), we run TPE on the related low-cost task such as optimizing \textit{MI} values, aiming to search for the SQL queries with high \textit{MI} values. Then we select top-k SQL queries (e.g., 50) with the highest \textit{MI} values, evaluate them, and use them to initialize the surrogate model (i.e. the KDEs) of the second round of TPE (Query-Generation Phase), which aims to search for the predicate-aware SQL query leads to the lowest validation loss of the ML model. 

\section{Query Template Identification}
In Section V, we introduce how to generate effective queries when the query template is fixed. In a practical scenario, users often do not know the detailed correlation between the training table and the relevant table. Thus, the attributes provided by users for forming predicates may not be the promising attribute combination for generating effective SQL queries. To further promise the generalization of \sqlgen, we try to identify the effective query template when users cannot provide explicit query templates.
In this section, we introduce the \textit{Query Template Identification} component that identifies promising query templates when users cannot provide explicit query templates.


\subsection{The Brute-Force Approach}
Assume $attr$ is a set of attributes from where we select a fixed attribute combination $P$ to construct a query template as Definition~\ref{def:query-template} described, then the possible number of query templates equals the number of subsets of $attr$, which is $2^{|attr|}$. The brute-force approach for identifying $n$ promising query templates is to calculate the effectiveness of all $2^{|attr|}$ query templates and select the query templates with $n$ highest effectiveness. Note that the cost of calculating the effectiveness of each query template $T \in \mathcal{S}_{attr}$ is different. That is because the size of $Q_T$ is different and the execution cost of each SQL query $q \in Q_T$ varies. Thus we denote $cost$ as the maximum cost of calculating the effectiveness of each query template $T \in \mathcal{S}_{attr}$. With the brute-force approach, the maximum cost of identifying promising query templates is $2^{|attr|}\cdot cost$. Because of the huge number of predicate-aware SQL queries in the query pool related to each query template, obviously, it is impractical to search for
global promising query templates with such an expensive cost.

\subsection{The Beam Search Approach}
To avoid the expensive calculation of the brute-force method and make the query template identification practical, inspired by \textit{Beam Search}~\cite{BeamSearch}, we map the search space of the query template into the tree-like search space, and employ the \textit{greedy} idea to explore the most promising part of a tree-structured search space in Figure~\ref{fig:query_template_space}. 

\subsubsection{The Tree-Structured Search Space.} 
\begin{figure}[t]\vspace{0em}
  \centering
  \includegraphics[width=\linewidth]{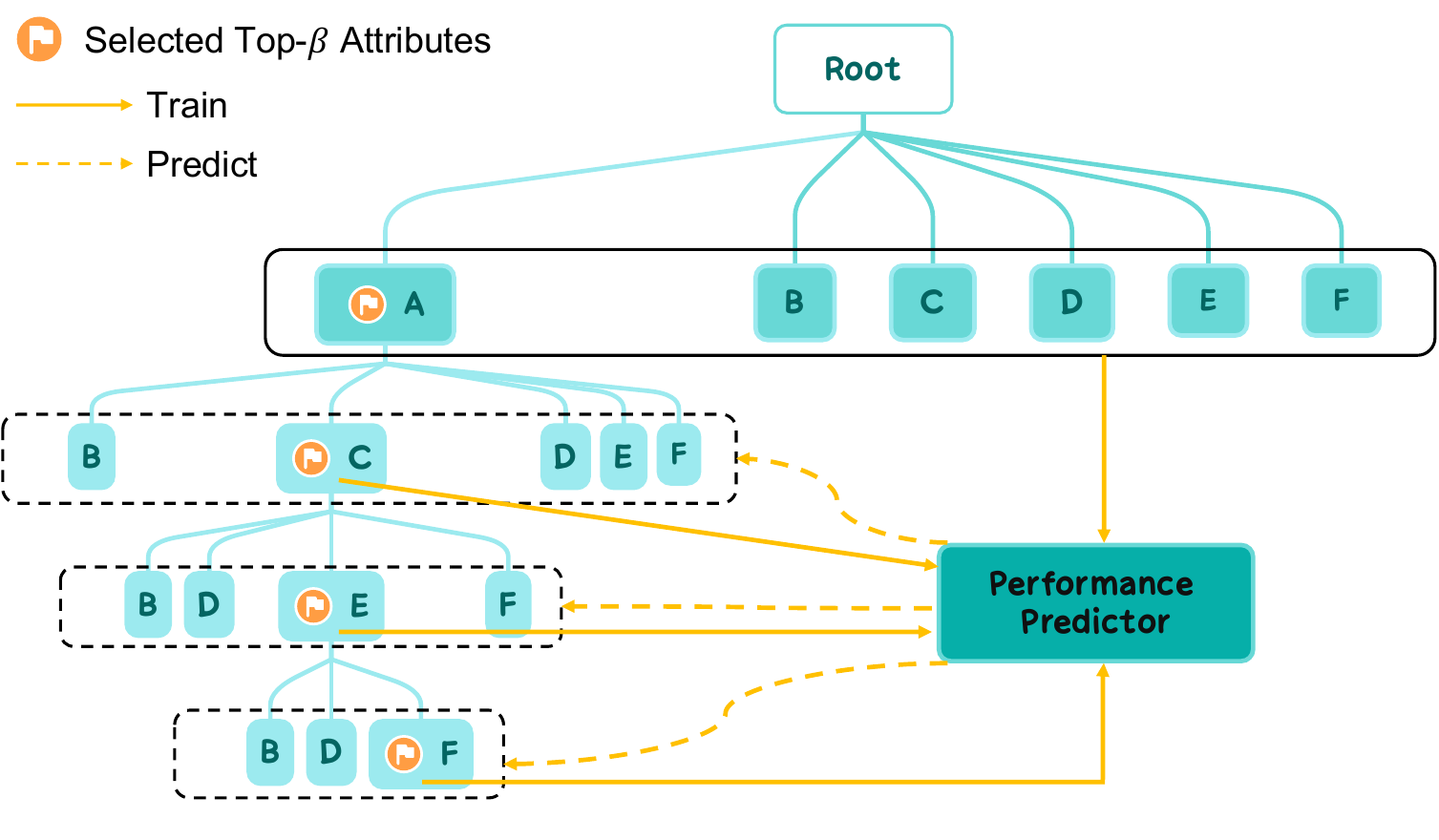}
  \caption{The illustration for the search space and the process of the Query Template Identification component. ($\beta = 1$)}
  \label{fig:query_template_space} \vspace{-1em}
\end{figure}

The different subsets of $attr$, i.e. attribute combinations, construct a tree-structured search space shown in Figure~\ref{fig:query_template_space}. Each node depicts one possible attribute combination, i.e. one possible query template. Nodes in the first layer indicate the attribute combinations formed by only one attribute (e.g. $\{A\}, \{B\}, \cdots$). Nodes in the second layer indicate the attribute combinations formed by two attributes (e.g. $\{A, B\}, \{A, C\}, \cdots$).
Obviously, the tree-structured search space in Figure~\ref{fig:query_template_space} expands exponentially. If the relevant table is high-dimensional, it is crucial to find the search direction smarter.  

\vspace{-.15em}



\subsubsection{The Identification Process with Beam Search.} The basic idea of \textit{Beam search}~\cite{BeamSearch} is to only expand the top-$\beta$ promising nodes in each layer. For example, Figure~\ref{fig:query_template_space} shows one typical expansion with $\beta=1$. Starting from the \texttt{Root} node, in the first layer, we get query templates formed by only one attribute (e.g. $\{A\}, \{B\}, \cdots$) and calculate their effectiveness. Then we pick up the top-1 node $\{A\}$ for the following expansion. 
In the second layer, we expand $\{A\}$ to $\{A, B\}, \{A, C\}, \cdots$, calculate their effectiveness and pick up the top-1 node to continue the expansion. The procedure is terminated when the max $depth$ for expansion is achieved. In Figure~\ref{fig:query_template_space}, we set the max $depth = 4$. 
Note that different query templates have different attribute combinations in the \texttt{WHERE} clauses. 
Thus, after the termination of the process in Figure ~\ref{fig:query_template_space}, we get 6+5+4+3=18 query templates and their effectiveness. The $n$ most promising query templates are identified by picking up query templates with $n$ highest effectiveness \textcolor{blue}{from all the 18 query templates}. The identification process with \textit{Beam Search} can decline the maximum cost of identifying promising query templates from $2^{|attr|}\cdot cost$ to $\Big(|attr| + \sum^{depth}_{i = 2}\beta\cdot (|attr| - i)\Big)\cdot cost$. For example, we calculate the effectiveness of 18 query templates in Figure ~\ref{fig:query_template_space} rather than $2^6=72$ query templates.



\subsection{Optimizations for The Identification Process}
However, the identification process described above is still not practical. That is because to get the top-$\beta$ promising nodes in each layer, all nodes in this layer should be evaluated. 
The evaluation result of a node, i.e. the effectiveness of a query template $T$ is determined by the most effective SQL query $q^* \in Q_T$ that minimizes the model validation loss. The optimal query $q^*$ can be identified through exhaustive enumeration of $Q_T$ by computing the actual validation loss associated with the ML model, which becomes computationally intensive when dealing with large training tables or complex models. To solve the above issue, we introduce a low-cost proxy to simulate the evaluation result of each node, i.e. the effectiveness of each query template. Instead of evaluating all nodes in each layer, we evaluate promising nodes by utilizing a performance predictor.  

\subsubsection{Optimization 1: Low-cost Proxy for Query Template Effectiveness.} 
Instead of calculating the real validation loss of the ML model, considering a low-cost proxy to simulate the real validation loss is more practical. 
To address this, we use the low-cost proxy like \textit{Mutual Information (MI)} to represent the real validation loss. MI is a well-established method in feature selection~\cite{MIFeatureSelect, brown2012conditional, guyon2003introduction}.
Given two random variables $X$ and $Y$, MI measures the dependency between the two variables. The higher MI value indicates higher dependency. 
Note that the effectiveness of query template $T$ equals the evaluation result of the most effective SQL query $q^* \in Q_T$, which leads to the lowest validation loss of the ML model compared to other SQL queries in $Q_T$. Thus, the \textit{MI} between the feature generated by $q^*$ and the labels can be the proxy of $T$'s effectiveness. 

Let us denote $cost_p$ as the maximum cost of calculating the low-cost proxy value for each query template $T \in \mathcal{S}_{attr}$. The low-cost proxy optimization can reduce the maximum cost of query template identification to $\Big(|attr| + \sum^{depth}_{i=2}\beta\cdot(|attr|-i)\Big)\cdot cost_p$. Obviously, it is much cheaper because $cost_p << cost$.

\subsubsection{Optimization 2: Promising Query Templates Prediction.} Even with the low-cost proxy, for selecting top-$\beta$ nodes in each layer, we still need to evaluate all nodes (i.e. all query templates) in this layer. Thus, it is essential to cut off unpromising nodes prior to prevent redundant evaluations. A predictor can be trained to predict whether a node, i.e. query templates can produce effective predicate-aware SQL queries or not. For training this predictor, we first need to encode query templates, then collect training data and do inference layer-by-layer. We introduce the details of encoding query templates and predicting whether the query templates are promising in the following content.

\begin{table}[t]
\vspace{0em}
\caption{Detailed information of datasets. "\# of Tables": the number of tables included in each dataset. "\# of rows in $R$": the number of rows in relevant tables. }
\vspace{0em}
\label{tab:dataset_statistic}
\centering
\small
\begin{adjustbox}{width=\linewidth}
\begin{tabular}{cccc}
\hline
{\color[HTML]{000000} \textbf{Dataset}}   & {\color[HTML]{000000} \textbf{\# of Tables}} & {\color[HTML]{000000} \textbf{\# of rows in R}} & {\color[HTML]{000000} \textbf{\# of Train/Valid/Test}} \\ \hline
\textbf{Tmall}                            & 3                                            & 6.5M                                            & 3.7w/1.2w/1.2w                                         \\
\textbf{Instacart}                        & 4                                            & 7.8M                                            & 3w/1w/1w                                               \\
\textbf{Student}                          & 2                                            & 1.6M                                            & 6k/2k/2k                                               \\
\textbf{Merchant}                         & 3                                            & 4.4M                                            & 3w/1w/1w                                               \\ \hline
\end{tabular}
\end{adjustbox}
\end{table}

\begin{itemize}
    \item \textbf{Encoding Query Templates.} The difference among query templates is the different attribute combinations in the \texttt{WHERE} clause. Thus we take one-hot encoding to encode query templates. Take the attributes in Figure~\ref{fig:query_template_space} as an example, there are six attributes \texttt{\{A, B, C, D, E, F\}} which can generate $2^6$ possible query templates. If the \texttt{WHERE} clause of a query template $T$ is composed by the attribute combination \texttt{\{A, C, E, F\}}, the encoding $e_{T}$=[1, 0, 1, 0, 1, 1]. 
    \item \textbf{Predicting Promising Query Templates.} As Figure~\ref{fig:query_template_space} shows, we train the predictor by collecting the training data layer-by-layer. In the first layer, we get query templates formed by only one attribute and the proxy values of them. Thus we can get 6 training data in the first layer and train the predictor. Before evaluating query templates in the second layer, we first use the trained predictor to predict the proxy value of each node, i.e. query template. Then pick up the top-$\beta$ query templates with top-$\beta$ highest prediction scores and calculate their proxy values.

    With the promising query template prediction, the maximum cost of query template identification can be finally reduced to $\Big(|attr| + \sum^{depth}_{i=2}\beta\Big)\cdot cost_p$, which is much cheaper than the brute-force approach and the original beam search approach.
\end{itemize}
\section{Experiments}\label{sec:exp}

We conduct extensive experiments using real-world datasets to evaluate \sqlgen. The experiments aim to answer main questions: 
(1) Can \sqlgen benefit from the proposed optimizations? (2) Can \sqlgen find more effective features compared to baselines on traditional ML models and deep models? (3) How does the performance of \sqlgen change when the important settings change?

\subsection{Experimental Settings}


\subsubsection{Datasets} We use the following \textcolor{blue}{6 datasets} including classification and regression tasks to conduct our experiments. The detailed information of the \textcolor{blue}{6 datasets} is shown in Table~\ref{tab:dataset_statistic}:

\begin{itemize}
    \item \textcolor{blue}{\textit{Covtype}~\cite{Covtype} aims to predict forest cover in four Colorado wilderness areas. This dataset includes only one table and we take itself as the relevant table.} 
    \item \textcolor{blue}{\textit{Household}~\cite{Household} aims to classify the families' poverty level by considering the observable household attributes. It contains only one table. We keep 5 features in the training table and put other 137 features into the relevant table.}
    \item \textit{Tmall}~\cite{Tmall} is a repeat buyer prediction data aiming to predict whether a customer will be a repeat buyer of a specific merchant. This dataset includes three tables and we join the user profile table and the user behaviour table into one relevant table. 
    \item \textit{Instacart}~\cite{Instacart} aims to predict whether a customer will purchase a commodity which has ``Banana" in its name. It contains four tables. We join the historical order table, the product table and the department table into one relevant table.
    \item \textit{Student}~\cite{Student} aims to use time series data generated by an online educational game to determine whether players will answer questions correctly. It contains two tables and we can directly consider the table containing the time series data as the relevant table. 
    \item \textit{Merchant}~\cite{Merchant} aims to recommend to users the merchant category they will buy in the next purchase. It contains three tables and we join the merchant information table and the historical transaction table into one relevant table.
\end{itemize}

\subsubsection{Detailed Information of Query Templates} 
Table~\ref{tab:search_space_detail} shows the aggregation functions ($F$) utilized by each dataset. It also shows the number of attributes for aggregation (\# of $A$) and the number of attributes in the relevant table that may be helpful for forming \texttt{WHERE} clause (\# of $attr$). The concrete names of attributes can be found in our technical report~\cite{FeatAug_Report}. The group-by keys ($K$) between the training and relevant table of each dataset are also shown in Table~\ref{tab:search_space_detail}. With these information, query templates and the related query pools can be constructed. 

\begin{table}[t]
\vspace{0em}
\caption{Detailed information of query templates.
"F": the aggregation functions. "\# of A": the number of attributes for aggregation. "\# of $attr$": the number of provided attributes that may be useful for forming \texttt{WHERE} clause. "K": the group-by keys between the training and relevant table. "\# of T": the number of query templates.}
\label{tab:search_space_detail}
\centering
\small
\begin{adjustbox}{width=\linewidth}
\begin{tabular}{c|cccc}
\hline
\textbf{Dataset}                                             & \textbf{Tmall}                                                                  & \textbf{Instacart}                & \textbf{Student}               & \textbf{Merchant}               \\ \hline
\textbf{F}                                                   & \multicolumn{4}{c}{\begin{tabular}[c]{@{}c@{}} \textsf{SUM, MIN, MAX, COUNT, AVG,} \\ \textsf{COUNT\_DISTINCT, VAR, VAR\_SAMPLE,}\\ \textsf{STD, STD\_SAMPLE, ENTROPY,}\\ \textsf{KURTOSIS, MODE, MAD, MEDIAN}\end{tabular}} \\ \hline
\textbf{\# of A}                                             & 6                                                                               & 6                                 & 10                             & 34                              \\ \hline
\textbf{\# of $attr$}                                             & 5                                                                               & 8                                 & 10                             & 15                              \\ \hline
\textbf{K}                                                   & \begin{tabular}[c]{@{}c@{}}\textsf{user\_id,}\\ \textsf{merchant\_id}\end{tabular}                & \textsf{user\_id}                          & \textsf{session\_id}                    & \textsf{merchant\_id}                    \\ \hline

\textbf{\# of T}                                             & $2^5$                                                                               & $2^8$                                 & $2^{10}$                            & $2^{15}$                           \\ \hline
\end{tabular}
\end{adjustbox}
\end{table}

\subsubsection{Baselines} 
We compare our \sqlgen with a variety of typical solutions. For all datasets, the first compared approach is \textit{Featuretools}~\cite{FeatureTools}. Note that \textit{Featuretools} cannot construct predicate-aware SQL queries and it does not filter out any useless SQL queries (i.e. features) during the generation process. \textcolor{blue}{Thus, we combine the SQL query (i.e. feature) generation process of \textit{Featuretools} with feature selectors and also consider them as the compared approaches. In this paper, we choose seven feature selectors by considering the feature selection approaches in Section II.} Another compared approach is the random approach, which randomly picks up query templates and predicate-aware SQL queries, i.e. features. \textcolor{blue}{For datasets with one-to-one relationship tables, two additional baselines dealing with this scenario, i.e. \textit{ARDA}~\cite{ARDA} and \textit{AutoFeature}~\cite{FARL} are also compared.}

\begin{itemize}
    \item \textbf{Featuretools} materalizes all features with \textit{Featuretools} without any feature selector. 
    \item \textbf{\textcolor{blue}{Featuretools + LR / GBDT Selector}} \textcolor{blue}{first generates features with \textit{Featuretools}, then uses these features to train a \textit{Logistic Regression} or a \textit{Gradient Boosting Decision Tree (GBDT)} classifier and selects the features with top feature importances.} 
    
    \item \textbf{\textcolor{blue}{Featuretools + MI / Chi2 / Gini Selector}} \textcolor{blue}{first generates features with \textit{Featuretools}, then uses \textit{Mutual Information (MI)} or \textit{Chi-square (Chi2)} or \textit{Gini index (Gini)} to evaluate the correlation between the features and the labels. Finally, the features with top correlations are selected. Note that \textit{Chi2} and \textit{Gini} are only suitable for classification tasks, and \textit{MI} is suitable for both classification and regression tasks} 

    \item \textbf{\textcolor{blue}{Featuretools + Forward Selector}} \textcolor{blue}{first generates features with \textit{Featuretools}. Then, in each iteration, the \textit{Forward Selector} adds the feature causing the highest improvement of the downstream ML model performance into the training table.} 

    \item \textbf{\textcolor{blue}{Featuretools + Backward Selector}} \textcolor{blue}{first generates features with \textit{Featuretools}. Then, in each iteration, the \textit{Backward Selector} removes the feature degrading the downstream ML model performance most.} 
    \item \textbf{\textcolor{blue}{ARDA}} \textcolor{blue}{heuristically uses a random injection-based feature augmentation to search good feature subsets from the relevant table. Note that ARDA works for datasets with one-to-one relationship tables.} 

    \item \textbf{\textcolor{blue}{AutoFeature}} \textcolor{blue}{is a RL-based automatic feature augmentation method working for datasets with one-to-one relationship tables. In each iteration, AutoFeature utilizes \textit{Multi-armed Bandit (MAB)} or \textit{Deep Q Network (DQN)} to predict the next action, i.e. the next feature to augment.} 
    
    \item \textbf{Random} first chooses query templates from the query template set randomly, then randomly searches predicate-aware SQL queries in each query pool of each query template. 
\end{itemize}

\textcolor{blue}{In our experiments, we utilize \textit{Featuretools} and \textit{Featuretools + Selectors} to generate 40 features. } 
We also use the random approach and \sqlgen to generate 40 predicate-aware SQL queries, (i.e. features) by selecting 8 query templates and 5 predicate-aware SQL queries in each query pool related to each query template. 

\subsubsection{ML Models} We evaluate our proposed method using three traditional ML models including \textit{Logistic Regression (LR), Random Forest (RF), XGBoost (XGB)} and one deep model \textit{DeepFM}~\cite{DeepFM}. We choose the three traditional ML models based on the recent survey~\cite{Kaggle-2021-survey}, which shows their effectiveness and popularity. \textit{LR} and \textit{RF} are the two most popular ML models. \textit{XGB} is a tree-based model which takes the first popularity of complex ML models. We choose  \textit{DeepFM} because it is effective and widely used in the industry, particularly for recommendation systems and advertising. Choosing \textit{DeepFM} as downstream ML models emphasizes the practical implications and potential benefits of \sqlgen in real-world applications.


\subsubsection{Metrics} \textcolor{blue}{For the classification datasets including \textit{Covtype} and \textit{Household}, we evaluate the performance with \textit{F1} score because they are multi-class datasets.} For the classification datasets including \textit{Tmall, Instacart} and \textit{Student}, we evaluate the performance with \textit{AUC}, where the receiver operator characteristic (ROC) is a probability curve displaying the performance over a series of thresholds and AUC is the area under the ROC curve. 
For the regression dataset \textit{Merchant}, we evaluate the performance using \textit{RMSE}.

\subsubsection{Implementation Details} For all the datasets, we set the ratio of train/valid/test as 0.6/0.2/0.2. We develop \sqlgen based on the TPE implementation in the Hyperopt library~\cite{Hyperopt}. The traditional ML models we used in experiments are constructed using Scikit-Learn library~\cite{scikit-learn}. The code is written in Python 3.8.10. 
Our experiments are conducted on one AWS EC2 r6idn.8xlarge instance (32 vCPUs and 256GB main memory) by default. All of the experiments are repeated five times and we report the average to avoid the influence of hardware, network and randomness.


\subsection{Can FeatAug Find More Effective Features for One-to-Many Relationship Tables?}

In this section, we compare \sqlgen with baselines to figure out whether \sqlgen can find more effective features. 

\begin{table}[t]
\caption{Overall performance of \sqlgen compared to baselines on datasets with one-to-many relationship tables. "FT": Featuretools without any feature selector. "FT+X": Featuretools with X Selector. For example, "FT+LR" means Featuretools + LR Selector.}
\label{tab:overall_performance}
\centering
\footnotesize
\begin{adjustbox}{width=\linewidth}
\begin{tabular}{cc|c|c|c|c}
\hline
\multicolumn{2}{c|}{{\color[HTML]{3531FF} \textbf{Dataset}}}                                                                 & {\color[HTML]{3531FF} \textbf{Tmall}}  & {\color[HTML]{3531FF} \textbf{Instacart}} & {\color[HTML]{3531FF} \textbf{Student}} & {\color[HTML]{3531FF} \textbf{Merchant}} \\ \hline
\multicolumn{2}{c|}{{\color[HTML]{3531FF} \textbf{Metric}}}                                                                  & {\color[HTML]{3531FF} AUC $\uparrow$}             & {\color[HTML]{3531FF} AUC $\uparrow$}                & {\color[HTML]{3531FF} AUC $\uparrow$}              & {\color[HTML]{3531FF} RMSE $\downarrow$}              \\ \hline
\multicolumn{1}{c|}{{\color[HTML]{3531FF} }}                                   & {\color[HTML]{3531FF} \textbf{FT}}          & {\color[HTML]{3531FF} 0.5610}          & {\color[HTML]{3531FF} 0.5679}             & {\color[HTML]{3531FF} 0.5269}           & {\color[HTML]{3531FF} 3.9677}            \\
\multicolumn{1}{c|}{{\color[HTML]{3531FF} }}                                   & {\color[HTML]{3531FF} \textbf{FT+LR}}       & {\color[HTML]{3531FF} 0.5641}          & {\color[HTML]{3531FF} 0.5877}             & {\color[HTML]{3531FF} 0.5427}           & {\color[HTML]{3531FF} 3.9914}            \\
\multicolumn{1}{c|}{{\color[HTML]{3531FF} }}                                   & {\color[HTML]{3531FF} \textbf{FT+GDBT}}     & {\color[HTML]{3531FF} 0.5620}          & {\color[HTML]{3531FF} 0.6100}             & {\color[HTML]{3531FF} 0.5003}           & {\color[HTML]{3531FF} 3.9678}            \\
\multicolumn{1}{c|}{{\color[HTML]{3531FF} }}                                   & {\color[HTML]{3531FF} \textbf{FT+MI}}       & {\color[HTML]{3531FF} 0.5550}          & {\color[HTML]{3531FF} 0.6054}             & {\color[HTML]{3531FF} 0.5061}           & {\color[HTML]{3531FF} 3.9670}            \\
\multicolumn{1}{c|}{{\color[HTML]{3531FF} }}                                   & {\color[HTML]{3531FF} \textbf{FT+Chi2}}     & {\color[HTML]{3531FF} 0.5620}          & {\color[HTML]{3531FF} 0.6002}             & {\color[HTML]{3531FF} 0.5450}           & {\color[HTML]{3531FF} -}                 \\
\multicolumn{1}{c|}{{\color[HTML]{3531FF} }}                                   & {\color[HTML]{3531FF} \textbf{FT+Gini}}     & {\color[HTML]{3531FF} 0.5626}          & {\color[HTML]{3531FF} 0.5746}             & {\color[HTML]{3531FF} 0.5846}           & {\color[HTML]{3531FF} -}                 \\
\multicolumn{1}{c|}{{\color[HTML]{3531FF} }}                                   & {\color[HTML]{3531FF} \textbf{FT+Forward}}  & {\color[HTML]{3531FF} 0.5580}          & {\color[HTML]{3531FF} 0.5877}             & {\color[HTML]{3531FF} 0.5756}           & {\color[HTML]{3531FF} 3.9735}            \\
\multicolumn{1}{c|}{{\color[HTML]{3531FF} }}                                   & {\color[HTML]{3531FF} \textbf{FT+Backward}} & {\color[HTML]{3531FF} 0.5554}          & {\color[HTML]{3531FF} 0.6027}             & {\color[HTML]{3531FF} 0.5500}           & {\color[HTML]{3531FF} 3.9699}            \\
\multicolumn{1}{c|}{{\color[HTML]{3531FF} }}                                   & {\color[HTML]{3531FF} \textbf{Random}}      & {\color[HTML]{3531FF} 0.5630}          & {\color[HTML]{3531FF} 0.6021}             & {\color[HTML]{3531FF} 0.5620}           & {\color[HTML]{3531FF} 3.9804}            \\
\multicolumn{1}{c|}{\multirow{-10}{*}{{\color[HTML]{3531FF} \textbf{LR}}}}     & {\color[HTML]{3531FF} \textbf{FeatAug}}     & {\color[HTML]{3531FF} \textbf{0.5749}} & {\color[HTML]{3531FF} \textbf{0.6369}}    & {\color[HTML]{3531FF} \textbf{0.5935}}  & {\color[HTML]{3531FF} \textbf{3.9538}}   \\ \hline
\multicolumn{1}{c|}{{\color[HTML]{3531FF} }}                                   & {\color[HTML]{3531FF} \textbf{FT}}          & {\color[HTML]{3531FF} 0.5568}          & {\color[HTML]{3531FF} 0.6349}             & {\color[HTML]{3531FF} 0.5730}           & {\color[HTML]{3531FF} 4.0752}            \\
\multicolumn{1}{c|}{{\color[HTML]{3531FF} }}                                   & {\color[HTML]{3531FF} \textbf{FT+LR}}       & {\color[HTML]{3531FF} 0.5526}          & {\color[HTML]{3531FF} 0.6311}             & {\color[HTML]{3531FF} 0.5649}           & {\color[HTML]{3531FF} 4.1576}            \\
\multicolumn{1}{c|}{{\color[HTML]{3531FF} }}                                   & {\color[HTML]{3531FF} \textbf{FT+GDBT}}     & {\color[HTML]{3531FF} 0.5494}          & {\color[HTML]{3531FF} 0.6488}             & {\color[HTML]{3531FF} 0.5736}           & {\color[HTML]{3531FF} 4.0637}            \\
\multicolumn{1}{c|}{{\color[HTML]{3531FF} }}                                   & {\color[HTML]{3531FF} \textbf{FT+MI}}       & {\color[HTML]{3531FF} 0.5548}          & {\color[HTML]{3531FF} 0.6507}             & {\color[HTML]{3531FF} 0.5778}           & {\color[HTML]{3531FF} 4.0882}            \\
\multicolumn{1}{c|}{{\color[HTML]{3531FF} }}                                   & {\color[HTML]{3531FF} \textbf{FT+Chi2}}     & {\color[HTML]{3531FF} 0.5367}          & {\color[HTML]{3531FF} 0.6343}             & {\color[HTML]{3531FF} \textbf{0.5821}}  & {\color[HTML]{3531FF} -}                 \\
\multicolumn{1}{c|}{{\color[HTML]{3531FF} }}                                   & {\color[HTML]{3531FF} \textbf{FT+Gini}}     & {\color[HTML]{3531FF} 0.5551}          & {\color[HTML]{3531FF} 0.6387}             & {\color[HTML]{3531FF} 0.5756}           & {\color[HTML]{3531FF} -}                 \\
\multicolumn{1}{c|}{{\color[HTML]{3531FF} }}                                   & {\color[HTML]{3531FF} \textbf{FT+Forward}}  & {\color[HTML]{3531FF} 0.5587}          & {\color[HTML]{3531FF} 0.6492}             & {\color[HTML]{3531FF} 0.5720}           & {\color[HTML]{3531FF} 4.1068}            \\
\multicolumn{1}{c|}{{\color[HTML]{3531FF} }}                                   & {\color[HTML]{3531FF} \textbf{FT+Backward}} & {\color[HTML]{3531FF} 0.5549}          & {\color[HTML]{3531FF} 0.6396}             & {\color[HTML]{3531FF} 0.5782}           & {\color[HTML]{3531FF} 4.0860}            \\
\multicolumn{1}{c|}{{\color[HTML]{3531FF} }}                                   & {\color[HTML]{3531FF} \textbf{Random}}      & {\color[HTML]{3531FF} 0.5848}          & {\color[HTML]{3531FF} 0.5830}             & {\color[HTML]{3531FF} 0.5575}           & {\color[HTML]{3531FF} 4.0161}            \\
\multicolumn{1}{c|}{\multirow{-10}{*}{{\color[HTML]{3531FF} \textbf{XGB}}}}    & {\color[HTML]{3531FF} \textbf{FeatAug}}     & {\color[HTML]{3531FF} \textbf{0.5898}} & {\color[HTML]{3531FF} \textbf{0.6844}}    & {\color[HTML]{3531FF} 0.5782}           & {\color[HTML]{3531FF} \textbf{4.0012}}   \\ \hline
\multicolumn{1}{c|}{{\color[HTML]{3531FF} }}                                   & {\color[HTML]{3531FF} \textbf{FT}}          & {\color[HTML]{3531FF} 0.5000}          & {\color[HTML]{3531FF} 0.5601}             & {\color[HTML]{3531FF} 0.5205}           & {\color[HTML]{3531FF} \textbf{4.0160}}   \\
\multicolumn{1}{c|}{{\color[HTML]{3531FF} }}                                   & {\color[HTML]{3531FF} \textbf{FT+LR}}       & {\color[HTML]{3531FF} 0.5010}          & {\color[HTML]{3531FF} 0.5675}             & {\color[HTML]{3531FF} 0.5178}           & {\color[HTML]{3531FF} 4.0778}            \\
\multicolumn{1}{c|}{{\color[HTML]{3531FF} }}                                   & {\color[HTML]{3531FF} \textbf{FT+GDBT}}     & {\color[HTML]{3531FF} 0.5000}          & {\color[HTML]{3531FF} 0.5723}             & {\color[HTML]{3531FF} 0.5262}           & {\color[HTML]{3531FF} 4.0274}            \\
\multicolumn{1}{c|}{{\color[HTML]{3531FF} }}                                   & {\color[HTML]{3531FF} \textbf{FT+MI}}       & {\color[HTML]{3531FF} 0.5028}          & {\color[HTML]{3531FF} 0.567}              & {\color[HTML]{3531FF} 0.5369}           & {\color[HTML]{3531FF} 4.0194}            \\
\multicolumn{1}{c|}{{\color[HTML]{3531FF} }}                                   & {\color[HTML]{3531FF} \textbf{FT+Chi2}}     & {\color[HTML]{3531FF} 0.5000}          & {\color[HTML]{3531FF} 0.5587}             & {\color[HTML]{3531FF} 0.5361}           & {\color[HTML]{3531FF} -}                 \\
\multicolumn{1}{c|}{{\color[HTML]{3531FF} }}                                   & {\color[HTML]{3531FF} \textbf{FT+Gini}}     & {\color[HTML]{3531FF} 0.5000}          & {\color[HTML]{3531FF} 0.5603}             & {\color[HTML]{3531FF} 0.5239}           & {\color[HTML]{3531FF} -}                 \\
\multicolumn{1}{c|}{{\color[HTML]{3531FF} }}                                   & {\color[HTML]{3531FF} \textbf{FT+Forward}}  & {\color[HTML]{3531FF} 0.5026}          & {\color[HTML]{3531FF} 0.5669}             & {\color[HTML]{3531FF} 0.5305}           & {\color[HTML]{3531FF} 4.0220}            \\
\multicolumn{1}{c|}{{\color[HTML]{3531FF} }}                                   & {\color[HTML]{3531FF} \textbf{FT+Backward}} & {\color[HTML]{3531FF} 0.5000}          & {\color[HTML]{3531FF} 0.5672}             & {\color[HTML]{3531FF} 0.5272}           & {\color[HTML]{3531FF} 4.0179}            \\
\multicolumn{1}{c|}{{\color[HTML]{3531FF} }}                                   & {\color[HTML]{3531FF} \textbf{Random}}      & {\color[HTML]{3531FF} 0.5572}          & {\color[HTML]{3531FF} 0.6057}             & {\color[HTML]{3531FF} 0.5432}           & {\color[HTML]{3531FF} 4.0246}            \\
\multicolumn{1}{c|}{\multirow{-10}{*}{{\color[HTML]{3531FF} \textbf{RF}}}}     & {\color[HTML]{3531FF} \textbf{FeatAug}}     & {\color[HTML]{3531FF} \textbf{0.5573}} & {\color[HTML]{3531FF} \textbf{0.6248}}    & {\color[HTML]{3531FF} \textbf{0.5636}}  & {\color[HTML]{3531FF} 4.0313}            \\ \hline
\multicolumn{1}{c|}{{\color[HTML]{3531FF} }}                                   & {\color[HTML]{3531FF} \textbf{FT}}          & {\color[HTML]{3531FF} 0.5818}          & {\color[HTML]{3531FF} 0.7001}             & {\color[HTML]{3531FF} 0.5685}           & {\color[HTML]{3531FF} 3.9840}            \\
\multicolumn{1}{c|}{{\color[HTML]{3531FF} }}                                   & {\color[HTML]{3531FF} \textbf{FT+LR}}       & {\color[HTML]{3531FF} 0.5970}          & {\color[HTML]{3531FF} 0.6988}             & {\color[HTML]{3531FF} 0.5824}           & {\color[HTML]{3531FF} 3.9925}            \\
\multicolumn{1}{c|}{{\color[HTML]{3531FF} }}                                   & {\color[HTML]{3531FF} \textbf{FT+GDBT}}     & {\color[HTML]{3531FF} 0.6074}          & {\color[HTML]{3531FF} 0.7085}             & {\color[HTML]{3531FF} 0.5967}           & {\color[HTML]{3531FF} 3.9327}            \\
\multicolumn{1}{c|}{{\color[HTML]{3531FF} }}                                   & {\color[HTML]{3531FF} \textbf{FT+MI}}       & {\color[HTML]{3531FF} 0.592}           & {\color[HTML]{3531FF} 0.7130}             & {\color[HTML]{3531FF} 0.595}            & {\color[HTML]{3531FF} 3.9656}            \\
\multicolumn{1}{c|}{{\color[HTML]{3531FF} }}                                   & {\color[HTML]{3531FF} \textbf{FT+Chi2}}     & {\color[HTML]{3531FF} 0.5878}          & {\color[HTML]{3531FF} 0.6974}             & {\color[HTML]{3531FF} 0.5773}           & {\color[HTML]{3531FF} -}                 \\
\multicolumn{1}{c|}{{\color[HTML]{3531FF} }}                                   & {\color[HTML]{3531FF} \textbf{FT+Gini}}     & {\color[HTML]{3531FF} 0.5914}          & {\color[HTML]{3531FF} 0.7092}             & {\color[HTML]{3531FF} 0.5967}           & {\color[HTML]{3531FF} -}                 \\
\multicolumn{1}{c|}{{\color[HTML]{3531FF} }}                                   & {\color[HTML]{3531FF} \textbf{FT+Forward}}  & {\color[HTML]{3531FF} 0.5717}          & {\color[HTML]{3531FF} 0.7021}             & {\color[HTML]{3531FF} 0.5945}           & {\color[HTML]{3531FF} 3.9863}            \\
\multicolumn{1}{c|}{{\color[HTML]{3531FF} }}                                   & {\color[HTML]{3531FF} \textbf{FT+Backward}} & {\color[HTML]{3531FF} 0.5828}          & {\color[HTML]{3531FF} 0.7047}             & {\color[HTML]{3531FF} 0.5923}           & {\color[HTML]{3531FF} 3.9649}            \\
\multicolumn{1}{c|}{{\color[HTML]{3531FF} }}                                   & {\color[HTML]{3531FF} \textbf{Random}}      & {\color[HTML]{3531FF} 0.5976}          & {\color[HTML]{3531FF} 0.6449}             & {\color[HTML]{3531FF} 0.6115}           & {\color[HTML]{3531FF} 3.9817}            \\
\multicolumn{1}{c|}{\multirow{-10}{*}{{\color[HTML]{3531FF} \textbf{DeepFM}}}} & {\color[HTML]{3531FF} \textbf{FeatAug}}     & {\color[HTML]{3531FF} \textbf{0.6226}} & {\color[HTML]{3531FF} \textbf{0.7364}}    & {\color[HTML]{3531FF} \textbf{0.6438}}  & {\color[HTML]{3531FF} \textbf{3.9277}}   \\ \hline
\end{tabular}
\end{adjustbox}
\end{table}



 \textcolor{blue}{We evaluate the performance of generated features on four datasets with one-to-many relationship tables in Table~\ref{tab:dataset_statistic} with four ML models, i.e. 16 scenarios in total.The effectiveness of \sqlgen is summarized in Table~\ref{tab:overall_performance}, highlighting its superiority over baselines in 14 scenarios with a maximum AUC increase of 10.14\% for classification tasks and a maximum RMSE reduction of 0.0740 for regression tasks. This demonstrates \sqlgen's broad applicability across both traditional ML and deep learning models. 
Note that \textit{Featuretools} generates features by constructing all possible SQL queries without considering \texttt{WHERE} clause, while \sqlgen generates features by considering both SQL queries with and without \texttt{WHERE} clause. If the predicate-aware SQL queries are useless, the performance of \sqlgen would be approximate to or worse than \textit{Featuretools}. However, \sqlgen shows performance improvement in most scenarios compared to \textit{Featuretools}.} 

\textcolor{blue}{It is notable that \sqlgen outperforms \textit{Random} in classification tasks, achieving average AUC improvements of 1.07\% on the \textit{Tmall} dataset, 6.17\% on the \textit{Instacart} dataset, and 2.62\% on the \textit{Student} dataset across various ML models. For regression tasks, it recorded an average RMSE improvement of 0.1848 on the \textit{Merchant} dataset across various ML models. These results demonstrate the effectiveness of Bayesian Optimization in identifying more efficient predicate-aware SQL queries compared to random search.}

\subsection{\textcolor{blue}{How does FeatAug perform for Single Table and One-to-One Relationship Tables?}}
\textcolor{blue}{
In this section, we extend the evaluation of \sqlgen's effectiveness to datasets with single table and one-to-one relationship tables. Note that the datasets with single table can also be transferred to the scenario with one-to-one relationship tables by duplicating itself as the relevant table. }

\textcolor{blue}{For the scenario with single table, we choose \textit{Covtype}~\cite{Covtype} dataset from UCI Machine Learning Repository~\cite{UCI}, which is used in the single table feature augmentation works~\cite{AutoFP, OpenFE}. \textit{Covtype}~\cite{Covtype} aims to predict forest cover in four Colorado wilderness areas. This dataset includes only one table and we take itself as the relevant table. For the scenario with one-to-one relationship tables, we choose \textit{Household}~\cite{Household} dataset which is been used by \textit{Featuretools} to show their demo~\cite{featuretoolsdemo, householddemo}. \textit{Household}~\cite{Household} aims to classify the families' poverty level by considering the observable household attributes. It contains only one table. We keep 5 features in the training table and put other 137 features into the relevant table.
The statistical information of these two datasets are shown in Table~\ref{tab:dataset_statistic_onetoone} and detailed information of query templates of these two datasets are shown in Table~\ref{tab:search_space_detail_onetoone}. }
\begin{table}[t]
\vspace{0em}
\caption{Detailed information of \textit{Covtype} and \textit{Household} datasets.}
\vspace{0em}
\label{tab:dataset_statistic_onetoone}
\centering
\small
\begin{adjustbox}{width=\linewidth}
\begin{tabular}{cccc}
\hline
{\color[HTML]{3531FF} \textbf{Dataset}}   & {\color[HTML]{3531FF} \textbf{\# of Tables}} & {\color[HTML]{3531FF} \textbf{\# of rows in R}} & {\color[HTML]{3531FF} \textbf{\# of Train/Valid/Test}} \\ \hline
{\color[HTML]{3531FF} \textbf{Covtype}}   & {\color[HTML]{3531FF} 1}                     & {\color[HTML]{3531FF} 50K}                      & {\color[HTML]{3531FF} 3w/1w/1w}                        \\
{\color[HTML]{3531FF} \textbf{Household}} & {\color[HTML]{3531FF} 1}                     & {\color[HTML]{3531FF} 9.5K}                     & {\color[HTML]{3531FF} 5.7k/1.9k/1.9k}                  \\\hline
\end{tabular}
\end{adjustbox}
\end{table}

\begin{table}[t]
\vspace{0em}
\caption{Detailed information of query templates of \textit{Covtype} and \textit{Household} datasets.}
\label{tab:search_space_detail_onetoone}
\centering
\small
\begin{adjustbox}{width=0.8\linewidth}
\begin{tabular}{c|cc}
\hline
{\color[HTML]{3531FF} \textbf{Dataset}}    & {\color[HTML]{3531FF} \textbf{Covtype}}                                                              & {\color[HTML]{3531FF} \textbf{Household}}                                                           \\ \hline
{\color[HTML]{3531FF} \textbf{F}}          & \multicolumn{2}{c}{{\color[HTML]{3531FF} \begin{tabular}[c]{@{}c@{}}\textsf{SUM, MIN, MAX, COUNT, AVG,}\\ \textsf{COUNT DISTINCT, VAR, VAR SAMPLE,}\\ \textsf{STD, STD SAMPLE, ENTROPY,}\\ \textsf{KURTOSIS, MODE, MAD, MEDIAN}\end{tabular}}} \\ \hline
{\color[HTML]{3531FF} \textbf{\# of A}}    & {\color[HTML]{3531FF} 54}                                                                            & {\color[HTML]{3531FF} 123}                                                                          \\ \hline
{\color[HTML]{3531FF} \textbf{\# of $attr$}} & {\color[HTML]{3531FF} 10}                                                                            & {\color[HTML]{3531FF} 20}                                                                           \\ \hline
{\color[HTML]{3531FF} \textbf{K}}          & {\color[HTML]{3531FF} \textsf{data\_index}}                                                                   & {\color[HTML]{3531FF} \textsf{data\_index}}                                                                  \\ \hline
{\color[HTML]{3531FF} \textbf{\# of T}}    & {\color[HTML]{3531FF} $2^{10}$}                                                     & {\color[HTML]{3531FF} $2^{20}$}                                                    \\ \hline
\end{tabular}
\end{adjustbox}
\end{table}

\textcolor{blue}{For datasets with single table and one-to-one relationship tables, two additional baselines dealing with these scenarios, i.e. \textit{ARDA}~\cite{ARDA} and \textit{AutoFeature}~\cite{FARL} are also compared:}
\begin{itemize}
    \item \textbf{\textcolor{blue}{ARDA}} \textcolor{blue}{heuristically uses a random injection-based feature augmentation to search good feature subsets from the relevant table. Note that ARDA works for datasets with one-to-one relationship tables.} 

    \item \textbf{\textcolor{blue}{AutoFeature}} \textcolor{blue}{is a RL-based automatic feature augmentation method working for datasets with one-to-one relationship tables. In each iteration, AutoFeature utilizes \textit{Multi-armed Bandit (MAB)} or \textit{Deep Q Network (DQN)} to predict the next action, i.e. the next feature to augment.} 
\end{itemize}

\textcolor{blue}{We evaluate the performance of generated features on two datasets in Table~\ref{tab:dataset_statistic_onetoone} with three ML models, i.e. 6 scenarios in total. That is because the two datasets are multi-class datasets and \textit{DeepFM} only works for binary classification tasks. The result of the effectiveness experiment is shown in Table~\ref{tab:overall_performance_onetoone}. We can see that \sqlgen outperforms all baselines in 4 scenarios. The \textit{F1} score improvement shows that \sqlgen can also work correctly and well for the datasets with single table and the one-to-one relationship tables. }

\begin{table}[t]
\caption{Overall performance of \sqlgen compared to baselines on dataset with one-to-one relationship tables. "FT+": Featuretools with the selector showing the highest performance.}
\label{tab:overall_performance_onetoone}
\centering
\footnotesize
\begin{tabular}{cc|c|c}
\hline
\multicolumn{2}{c|}{{\color[HTML]{3531FF} \textbf{Dataset}}}                                                               & {\color[HTML]{3531FF} \textbf{Covtype}} & {\color[HTML]{3531FF} \textbf{Household}} \\ \hline
\multicolumn{2}{c|}{{\color[HTML]{3531FF} \textbf{Metric}}}                                                                & {\color[HTML]{3531FF} F1 $\uparrow$}               & {\color[HTML]{3531FF} F1 $\uparrow$}                 \\ \hline
\multicolumn{1}{c|}{{\color[HTML]{3531FF} }}                                & {\color[HTML]{3531FF} \textbf{FT}}           & {\color[HTML]{3531FF} 0.1681}           & {\color[HTML]{3531FF} \textbf{0.2378}}    \\
\multicolumn{1}{c|}{{\color[HTML]{3531FF} }}                                & {\color[HTML]{3531FF} \textbf{FT+LR}}        & {\color[HTML]{3531FF} 0.1461}           & {\color[HTML]{3531FF} 0.1434}             \\
\multicolumn{1}{c|}{{\color[HTML]{3531FF} }}                                & {\color[HTML]{3531FF} \textbf{FT+GBDT}}      & {\color[HTML]{3531FF} 0.1248}           & {\color[HTML]{3531FF} 0.2356}             \\
\multicolumn{1}{c|}{{\color[HTML]{3531FF} }}                                & {\color[HTML]{3531FF} \textbf{FT+MI}}        & {\color[HTML]{3531FF} 0.1422}           & {\color[HTML]{3531FF} 0.2356}             \\
\multicolumn{1}{c|}{{\color[HTML]{3531FF} }}                                & {\color[HTML]{3531FF} \textbf{FT+Chi2}}      & {\color[HTML]{3531FF} 0.1514}           & {\color[HTML]{3531FF} 0.2302}             \\
\multicolumn{1}{c|}{{\color[HTML]{3531FF} }}                                & {\color[HTML]{3531FF} \textbf{FT+Gini}}      & {\color[HTML]{3531FF} 0.1559}           & {\color[HTML]{3531FF} 0.2356}             \\
\multicolumn{1}{c|}{{\color[HTML]{3531FF} }}                                & {\color[HTML]{3531FF} \textbf{FT+Forward}}   & {\color[HTML]{3531FF} -}                & {\color[HTML]{3531FF} -}                  \\
\multicolumn{1}{c|}{{\color[HTML]{3531FF} }}                                & {\color[HTML]{3531FF} \textbf{FT+Backward}}  & {\color[HTML]{3531FF} -}                & {\color[HTML]{3531FF} -}                  \\
\multicolumn{1}{c|}{{\color[HTML]{3531FF} }}                                & {\color[HTML]{3531FF} \textbf{ARDA}}         & {\color[HTML]{3531FF} 0.2275}           & {\color[HTML]{3531FF} 0.2020}             \\
\multicolumn{1}{c|}{{\color[HTML]{3531FF} }}                                & {\color[HTML]{3531FF} \textbf{AutoFeat-MAB}} & {\color[HTML]{3531FF} 0.2688}           & {\color[HTML]{3531FF} 0.1424}             \\
\multicolumn{1}{c|}{{\color[HTML]{3531FF} }}                                & {\color[HTML]{3531FF} \textbf{AutoFeat-DQN}} & {\color[HTML]{3531FF} 0.1930}           & {\color[HTML]{3531FF} 0.2161}             \\
\multicolumn{1}{c|}{{\color[HTML]{3531FF} }}                                & {\color[HTML]{3531FF} \textbf{Random}}       & {\color[HTML]{3531FF} 0.2942}           & {\color[HTML]{3531FF} 0.2112}             \\
\multicolumn{1}{c|}{\multirow{-13}{*}{{\color[HTML]{3531FF} \textbf{LR}}}}  & {\color[HTML]{3531FF} \textbf{FeatAug}}      & {\color[HTML]{3531FF} \textbf{0.3084}}  & {\color[HTML]{3531FF} 0.2159}             \\ \hline
\multicolumn{1}{c|}{{\color[HTML]{3531FF} }}                                & {\color[HTML]{3531FF} \textbf{FT}}           & {\color[HTML]{3531FF} 0.7582}           & {\color[HTML]{3531FF} 0.2718}             \\
\multicolumn{1}{c|}{{\color[HTML]{3531FF} }}                                & {\color[HTML]{3531FF} \textbf{FT+LR}}        & {\color[HTML]{3531FF} 0.3567}           & {\color[HTML]{3531FF} 0.2333}             \\
\multicolumn{1}{c|}{{\color[HTML]{3531FF} }}                                & {\color[HTML]{3531FF} \textbf{FT+GBDT}}      & {\color[HTML]{3531FF} 0.5067}           & {\color[HTML]{3531FF} 0.2782}             \\
\multicolumn{1}{c|}{{\color[HTML]{3531FF} }}                                & {\color[HTML]{3531FF} \textbf{FT+MI}}        & {\color[HTML]{3531FF} 0.5545}           & {\color[HTML]{3531FF} 0.2920}             \\
\multicolumn{1}{c|}{{\color[HTML]{3531FF} }}                                & {\color[HTML]{3531FF} \textbf{FT+Chi2}}      & {\color[HTML]{3531FF} 0.5981}           & {\color[HTML]{3531FF} 0.2903}             \\
\multicolumn{1}{c|}{{\color[HTML]{3531FF} }}                                & {\color[HTML]{3531FF} \textbf{FT+Gini}}      & {\color[HTML]{3531FF} 0.5368}           & {\color[HTML]{3531FF} 0.2839}             \\
\multicolumn{1}{c|}{{\color[HTML]{3531FF} }}                                & {\color[HTML]{3531FF} \textbf{FT+Forward}}   & {\color[HTML]{3531FF} -}                & {\color[HTML]{3531FF} -}                  \\
\multicolumn{1}{c|}{{\color[HTML]{3531FF} }}                                & {\color[HTML]{3531FF} \textbf{FT+Backward}}  & {\color[HTML]{3531FF} -}                & {\color[HTML]{3531FF} -}                  \\
\multicolumn{1}{c|}{{\color[HTML]{3531FF} }}                                & {\color[HTML]{3531FF} \textbf{ARDA}}         & {\color[HTML]{3531FF} 0.6422}           & {\color[HTML]{3531FF} 0.2735}             \\
\multicolumn{1}{c|}{{\color[HTML]{3531FF} }}                                & {\color[HTML]{3531FF} \textbf{AutoFeat-MAB}} & {\color[HTML]{3531FF} 0.7766}           & {\color[HTML]{3531FF} 0.2927}             \\
\multicolumn{1}{c|}{{\color[HTML]{3531FF} }}                                & {\color[HTML]{3531FF} \textbf{AutoFeat-DQN}} & {\color[HTML]{3531FF} 0.7766}           & {\color[HTML]{3531FF} 0.2453}             \\
\multicolumn{1}{c|}{{\color[HTML]{3531FF} }}                                & {\color[HTML]{3531FF} \textbf{Random}}       & {\color[HTML]{3531FF} \textbf{0.7800} }                 & {\color[HTML]{3531FF} 0.2666}             \\
\multicolumn{1}{c|}{\multirow{-13}{*}{{\color[HTML]{3531FF} \textbf{XGB}}}} & {\color[HTML]{3531FF} \textbf{FeatAug}}      & {\color[HTML]{3531FF} 0.7769}  & {\color[HTML]{3531FF} \textbf{0.3024}}    \\ \hline
\multicolumn{1}{c|}{{\color[HTML]{3531FF} }}                                & {\color[HTML]{3531FF} \textbf{FT}}           & {\color[HTML]{3531FF} 0.6289}           & {\color[HTML]{3531FF} 0.2444}             \\
\multicolumn{1}{c|}{{\color[HTML]{3531FF} }}                                & {\color[HTML]{3531FF} \textbf{FT+LR}}        & {\color[HTML]{3531FF} 0.3612}           & {\color[HTML]{3531FF} 0.2337}             \\
\multicolumn{1}{c|}{{\color[HTML]{3531FF} }}                                & {\color[HTML]{3531FF} \textbf{FT+GBDT}}      & {\color[HTML]{3531FF} 0.5201}           & {\color[HTML]{3531FF} 0.2534}             \\
\multicolumn{1}{c|}{{\color[HTML]{3531FF} }}                                & {\color[HTML]{3531FF} \textbf{FT+MI}}        & {\color[HTML]{3531FF} 0.5617}           & {\color[HTML]{3531FF} 0.2584}             \\
\multicolumn{1}{c|}{{\color[HTML]{3531FF} }}                                & {\color[HTML]{3531FF} \textbf{FT+Chi2}}      & {\color[HTML]{3531FF} 0.5611}           & {\color[HTML]{3531FF} 0.2522}             \\
\multicolumn{1}{c|}{{\color[HTML]{3531FF} }}                                & {\color[HTML]{3531FF} \textbf{FT+Gini}}      & {\color[HTML]{3531FF} 0.5434}           & {\color[HTML]{3531FF} 0.2526}             \\
\multicolumn{1}{c|}{{\color[HTML]{3531FF} }}                                & {\color[HTML]{3531FF} \textbf{FT+Forward}}   & {\color[HTML]{3531FF} -}                & {\color[HTML]{3531FF} -}                  \\
\multicolumn{1}{c|}{{\color[HTML]{3531FF} }}                                & {\color[HTML]{3531FF} \textbf{FT+Backward}}  & {\color[HTML]{3531FF} -}                & {\color[HTML]{3531FF} -}                  \\
\multicolumn{1}{c|}{{\color[HTML]{3531FF} }}                                & {\color[HTML]{3531FF} \textbf{ARDA}}         & {\color[HTML]{3531FF} 0.6573}           & {\color[HTML]{3531FF} 0.2639}             \\
\multicolumn{1}{c|}{{\color[HTML]{3531FF} }}                                & {\color[HTML]{3531FF} \textbf{AutoFeat-MAB}} & {\color[HTML]{3531FF} 0.7814}           & {\color[HTML]{3531FF} 0.2278}             \\
\multicolumn{1}{c|}{{\color[HTML]{3531FF} }}                                & {\color[HTML]{3531FF} \textbf{AutoFeat-DQN}} & {\color[HTML]{3531FF} 0.6884}           & {\color[HTML]{3531FF} 0.2371}             \\
\multicolumn{1}{c|}{{\color[HTML]{3531FF} }}                                & {\color[HTML]{3531FF} \textbf{Random}}       & {\color[HTML]{3531FF} 0.7964}                 & {\color[HTML]{3531FF} 0.2616}             \\
\multicolumn{1}{c|}{\multirow{-13}{*}{{\color[HTML]{3531FF} \textbf{RF}}}}  & {\color[HTML]{3531FF} \textbf{FeatAug}}      & {\color[HTML]{3531FF} \textbf{0.8074}}  & {\color[HTML]{3531FF} \textbf{0.3003}}    \\ \hline
\end{tabular}
\end{table}

\subsection{Can FeatAug Benefit from The Proposed Optimizations?}
In this section, we examine whether the proposed optimizations can benefit \sqlgen. The two main optimizations we proposed in this paper are the warm-up part in the \textit{SQL Generation} component and the \textit{Query Template Identification} component. The results are shown in Table~\ref{tab:ablation_study}. 

\subsubsection{Can \sqlgen Benefit from The Warm-up in SQL Generation?}
To study the benefit of the warm-up in the \textit{SQL Generation} component, we drop the warm-up part in the \textit{SQL Generation} component and Table~\ref{tab:ablation_study} shows the performance gap w/o the warm-up. In our implementation of \sqlgen, the process of the warm-up includes running TPE on the related low-cost task (i.e. optimizing MI value) for 200 iterations, selecting SQL queries with top-50 \textit{MI} values and evaluating them to initialize the surrogate model. Then we run 40 iterations of TPE with the warm-started surrogate model. For fair comparison, we do not simply drop the whole process above and run 40 iterations of TPE because the evaluating time of the top-50 SQL queries in the warm-up part cannot be neglected. Instead, we drop the warm-up part by only running 50+40=90 iterations of TPE. In most scenarios, the warm-up part can lead to better performance, which shows the effectiveness of transferring knowledge from the relevant tasks. An interesting observation is that the effectiveness of the warm-up part under different scenarios is different, which is highly related to datasets and the downstream ML models. Despite \textit{MI}, there are also other static characteristics like \textit{Spearman Correlation} are also alternatives of the proxy. We explore the effectiveness of different proxies in Section 7.4. Figuring out the proxy that contributes most to the warm-up prior is an interesting direction for future research.

\begin{table}[t]
\caption{Ablation study of \sqlgen. "NoQTI": \sqlgen without the Query Template Identification component. "NoWU": \sqlgen without the warm-up part in the SQL Query Generation component. "Full": \sqlgen with both the warm-up part in the in the SQL Query Generation component and the Query Template Identification component.}
\label{tab:ablation_study}
\centering
\footnotesize
\begin{adjustbox}{width=\linewidth}
\begin{tabular}{cc|ccc|c}
\hline
\multicolumn{2}{c|}{\textbf{Dataset}}                                                                                      & \textbf{Tmall}  & \textbf{Instacart} & \textbf{Student} & \textbf{Merchant} \\ \hline
\multicolumn{2}{c|}{\textbf{Metric}}                                                                                       & AUC $\uparrow$             & AUC $\uparrow$                & AUC $\uparrow$              & RMSE $\downarrow$              \\ \hline
\multicolumn{1}{c|}{\multirow{3}{*}{\textbf{LR}}}     & \textbf{\begin{tabular}[c]{@{}c@{}}FeatAug\\ (NoQTI)\end{tabular}} & 0.5257          & 0.5000             & 0.5000           & 3.9855            \\ \cline{2-6} 
\multicolumn{1}{c|}{}                                 & \textbf{\begin{tabular}[c]{@{}c@{}}FeatAug\\ (NoWU)\end{tabular}}  & 0.5650          & 0.6354             & \textbf{0.5935}  & 3.9549            \\ \cline{2-6} 
\multicolumn{1}{c|}{}                                 & \textbf{\begin{tabular}[c]{@{}c@{}}FeatAug\\ (Full)\end{tabular}}  & \textbf{0.5749} & \textbf{0.6369}    & \textbf{0.5935}  & \textbf{3.9538}   \\ \hline
\multicolumn{1}{c|}{\multirow{3}{*}{\textbf{XGB}}}    & \textbf{\begin{tabular}[c]{@{}c@{}}FeatAug\\ (NoQTI)\end{tabular}} & 0.5331          & 0.5000             & 0.5000           & 4.0176            \\ \cline{2-6} 
\multicolumn{1}{c|}{}                                 & \textbf{\begin{tabular}[c]{@{}c@{}}FeatAug\\ (NoWU)\end{tabular}}  & 0.5812          & 0.6794             & \textbf{0.5782}  & 4.0084            \\ \cline{2-6} 
\multicolumn{1}{c|}{}                                 & \textbf{\begin{tabular}[c]{@{}c@{}}FeatAug\\ (Full)\end{tabular}}  & \textbf{0.5898} & \textbf{0.6844}    & \textbf{0.5782}  & \textbf{4.0012}   \\ \hline
\multicolumn{1}{c|}{\multirow{3}{*}{\textbf{RF}}}     & \textbf{\begin{tabular}[c]{@{}c@{}}FeatAug\\ (NoQTI)\end{tabular}} & 0.5325          & 0.5000             & 0.5000           & \textbf{4.0063}   \\ \cline{2-6} 
\multicolumn{1}{c|}{}                                 & \textbf{\begin{tabular}[c]{@{}c@{}}FeatAug\\ (NoWU)\end{tabular}}  & 0.5526          & 0.6063             & 0.5582           & 4.0567            \\ \cline{2-6} 
\multicolumn{1}{c|}{}                                 & \textbf{\begin{tabular}[c]{@{}c@{}}FeatAug\\ (Full)\end{tabular}}  & \textbf{0.5573} & \textbf{0.6248}    & \textbf{0.5636}  & 4.0313            \\ \hline
\multicolumn{1}{c|}{\multirow{3}{*}{\textbf{DeepFM}}} & \textbf{\begin{tabular}[c]{@{}c@{}}FeatAug\\ (NoQTI)\end{tabular}} & 0.5284          & 0.5000             & 0.5000           & 3.9942            \\ \cline{2-6} 
\multicolumn{1}{c|}{}                                 & \textbf{\begin{tabular}[c]{@{}c@{}}FeatAug\\ (NoWU)\end{tabular}}  & 0.6186          & 0.7330             & 0.6303           & 3.9398            \\ \cline{2-6} 
\multicolumn{1}{c|}{}                                 & \textbf{\begin{tabular}[c]{@{}c@{}}FeatAug\\ (Full)\end{tabular}}  & \textbf{0.6226} & \textbf{0.7364}    & \textbf{0.6438}  & \textbf{3.9277}   \\ \hline
\end{tabular}
\end{adjustbox}
\end{table}

\subsubsection{Can \sqlgen Benefit from Query Template Identification?} To study the benefit of the query template identification component, we drop the query template identification part in \sqlgen and Table~\ref{tab:ablation_study} also shows the performance gap w/o the query template identification. Recall that for \texttt{Tmall}, \texttt{Instacart} and \texttt{Merchant} dateset, we provide a set of attributes which may be useful for forming the \texttt{WHERE} clause, while for \texttt{Student} dataset, we directly consider all the attributes in the relevant table. For dropping the \textit{Query Template Identification} component, we take the same attribute sets for each dataset to construct a query template and search for effective SQL query in the related query pool. In 15 out of 16 scenarios, adding the query template identification component can improve the performance significantly, which shows the effectiveness of picking up the promising query templates. Without the query template identification, there is only one possible query template constructed by the set of attributes provided by users, which leads to an unpromising query pool.

\subsubsection{Can the Query Template Identification Component Benefit from the Two Optimizations?}
\begin{figure*}[t]
  \vspace{-1em}
  \centering
  \includegraphics[width=\linewidth]{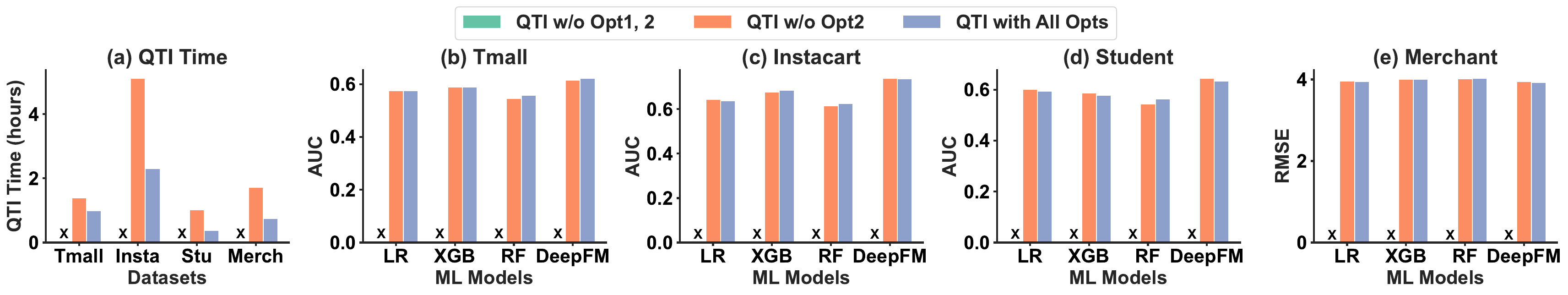}
  \caption{The ablation study of two optimization in the Query Template Identification component. (a): the running time of the Query Template Identification component w/o the two optimizations. ``X" means that the program cannot complete in 6 hours. (b) - (e): the performance comparison among \sqlgen with different Query Template Identification components.}
  \label{fig:qti-ablation} 
\end{figure*}

To study whether the two optimizations in Section VI can indeed speed up the Query Template Identification Component, we drop each optimization and Figure~\ref{fig:qti-ablation} (a) shows the running time w/o the two optimizations. Without any optimizations, the initial Beam Search cannot complete query template identification in 6 hours on any dataset, which indicates that the initial Beam Search is very time-consuming. With only \textit{Low-cost Proxy Optimization}, the Query Template Identification component can be completed in 2 hours on \textit{Tmall, Student} and \textit{Merchant} datasets, and in 5 hours on \textit{Instacart} dataset, which already speeds up the initial Beam Search. With all optimizations, the Query Template Identification component can be speed up \textbf{1.4x - 2.8x} compared to the method with only \textit{Low-cost Proxy Optimization}. We also explore whether adding the two optimizations will hurt the performance of \sqlgen severely.  Figure~\ref{fig:qti-ablation} (b) - (e) shows the comparison results. For all the four datasets and all the four downstreaming ML models (in total 16 scenarios), adding the \textit{Promising Query Template Prediction Optimization} hurts little performance of \sqlgen. The comparison results reveals that the \textit{Promising Query Template Prediction Optimization} can cut off unpromising query templates prior precisely and fast, and keep promising query templates to produce effective predicate-awared SQL queries. 

\subsection{In-Depth Analysis of FeatAug}

In this section, we perform the in-depth analysis of the performance impact to \sqlgen under different settings. 

\begin{table*}[t]
\caption{Performance of \sqlgen by varying the low-cost proxy. "SC" takes the Spearman Correlation as the low-cost proxy. "MI" takes the Mutual Information as the low-cost proxy. "LR" takes the the Logistic Regression model as the low-cost proxy.}
\label{tab:compare_proxies}
\centering
\footnotesize
\begin{adjustbox}{width=\linewidth}
\begin{tabular}{c|c|ccc|ccc|ccc|ccc}
\hline
\multirow{2}{*}{\textbf{Dataset}} & \multirow{2}{*}{\textbf{Metric}} & \multicolumn{3}{c|}{\textbf{LR}}                    & \multicolumn{3}{c|}{\textbf{XGB}}               & \multicolumn{3}{c|}{\textbf{RF}}                    & \multicolumn{3}{c}{\textbf{DeepFM}}         \\ \cline{3-14} 
                                  &                                  & \textbf{SC}     & \textbf{MI}     & \textbf{LR}     & \textbf{SC}     & \textbf{MI}     & \textbf{LR} & \textbf{SC}     & \textbf{MI}     & \textbf{LR}     & \textbf{SC} & \textbf{MI}     & \textbf{LR} \\ \hline
\textbf{Tmall}                    & AUC $\uparrow$                              & 0.5629          & \textbf{0.5749} & 0.5537          & 0.5854          & \textbf{0.5898} & 0.5888      & 0.5549          & \textbf{0.5573} & 0.5396          & 0.6177      & \textbf{0.6226} & 0.6135      \\ \cline{1-2}
\textbf{Instacart}                & AUC $\uparrow$                              & 0.6168          & 0.6369          & \textbf{0.6476} & 0.6632          & \textbf{0.6844} & 0.6057      & 0.6086          & 0.6248          & \textbf{0.6670} & 0.7266      & \textbf{0.7364} & 0.7269      \\ \cline{1-2}
\textbf{Student}                  & AUC $\uparrow$                              & \textbf{0.5935} & \textbf{0.5935} & 0.5846          & 0.5772          & \textbf{0.5782} & 0.5517      & 0.5687          & 0.5636          & \textbf{0.5750} & 0.6396      & \textbf{0.6438} & 0.6382      \\ \cline{1-2}
\textbf{Merchant}                 & RMSE $\downarrow$                             & 3.9623          & 3.9538          & \textbf{3.9756} & \textbf{3.9943} & 4.0012          & 4.0053      & \textbf{4.0230} & 4.0313          & 4.0666          & 3.9464      & \textbf{3.9277} & 3.9799      \\ \hline
\end{tabular}
\end{adjustbox}
\end{table*}

\subsubsection{Varying Number of Query Template.} 
\begin{figure}[t]
  \centering
  \includegraphics[width=\linewidth]{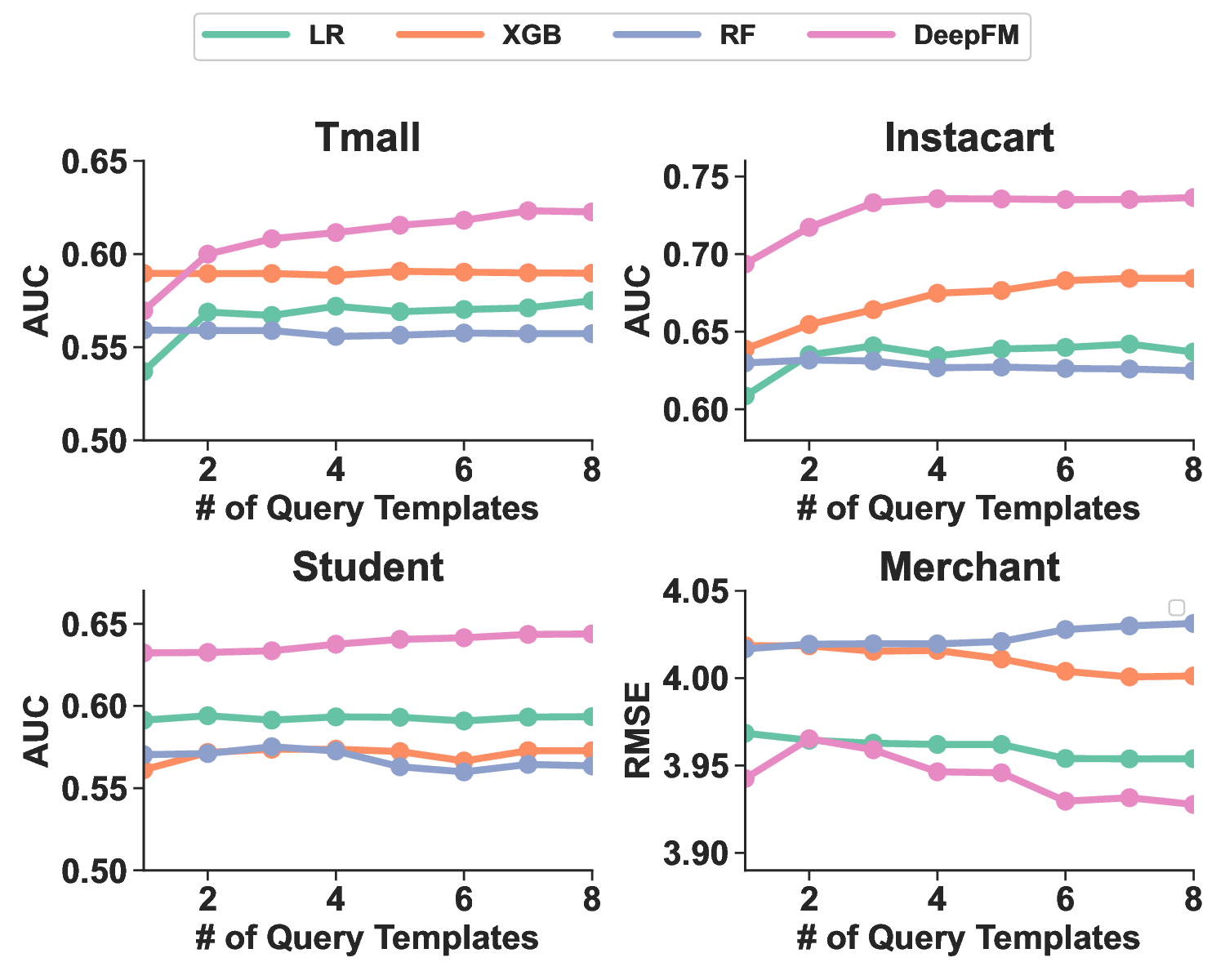}
  \caption{The trend of performance by varying the number of query templates. }
  \label{fig:vary_query_template} 
\end{figure}

Recall that when utilizing \sqlgen to generate effective SQL queries, we pick up 8 promising query templates and search for 5 effective SQL queries in each query pool. It is interesting whether more query templates cause better performance. Figure~\ref{fig:vary_query_template} shows the trend of performance by varying the number of query templates. We show all the trends on our 4 datasets and 4 downstream ML models. We have three interesting observations. 

 Firstly, in most cases (9 out of 16 scenarios), the increase in the number of query templates brings performance improvement to downstream ML models. It shows that considering multiple query templates is more helpful than only a single query template, which matches what data scientists really do in practice. Secondly, there is no fixed number of query templates that fit all scenarios. For \textit{Tmall} dataset with DeepFM model, \sqlgen converges when the number of query templates is 7, while \sqlgen converges at 3 when the dataset is \textit{Instacart} with DeepFM model. Thirdly, the deep model i.e. \textit{DeepFM} can get benefits easily from the increased number of query templates, while traditional ML models including \textit{LR, XGB} and \textit{RF} keep stable in most cases even the number of query templates increases. That is because the deep models can perform feature interaction automatically, and the increase of the number of query templates provides more opportunity for deep models to synthesize generated features into more informative features.

\subsubsection{Varying The Low-cost Proxy.} We explore the sensitivity of \sqlgen by varying the low-cost proxy and recommend the low-cost proxy in practical scenarios. We consider three low-cost proxies including:
\begin{itemize}
    \item \textit{Spearman's Correlation (SC)}: Given two variables $X$ and $Y$, Spearman's Correlation measures the strength and direction of the monotonic relationship between them. 
    It is defined as the Pearson correlation coefficient between the rank values of $X$ and $Y$, which is formulated as:
    \begin{align*}
        \rho = 1 - \frac{6\sum d_{i}^{2}}{n(n^{2} - 1)}
    \end{align*}
    where $d_{i}$ is the difference between the ranks of corresponding values of $X$ and $Y$, and $n$ is the number of observations. A higher absolute value of $\rho$ (close to 1 or -1) indicates a stronger monotonic dependency.
    
    \item \textit{Mutual Information (MI)}: Given two random variables $X$ and $Y$, MI measures the dependency between the two variables.
    which is defined as 
    \begin{align*}
        I(X;Y) = H(X) - H(X | Y)
    \end{align*}
    where $H(X)$ is the entropy of $X$ and $H(X | Y)$ is the conditional entropy for $X$ given $Y$. The higher MI value indicates higher dependency. 
    \item \textit{Logistic Regression (LR)}: \textit{LR} takes the performance of \textit{LR} model as the proxy of other ML models.
\end{itemize}

As we can see in Table~\ref{tab:compare_proxies}, \textit{SC} performs best in 2 out of 16 scenarios, \textit{LR} performs best in 3 out of 16 scenarios, and \textit{MI} is the most effective proxy in most cases, i.e. 11 out of 16 scenarios. The result suggests the entropies calculated in \textit{MI} can simulate the performance of ML models well in both classification and regression tasks. Surprisingly, \textit{SC} is competitive to \textit{MI} in 10 out of 16 scenarios. Note that the AUC score represents the probability that a classifier will rank a randomly chosen positive instance higher than a randomly chosen negative one. Thus the monotonic dependency that \textit{SC} measures is helpful for getting higher AUC score. The RMSE score can also benefit from \textit{SC}.
However, \textit{LR} proxy is not competitive with \textit{LR} and \textit{SC} in most cases, i.e. 10 out of 16 scenarios, suggesting that the performance of \textit{LR} cannot represent the performance of other ML models well.

\subsection{Scalability Analysis of FeatAug}
 As described in Section IV, the \sqlgen framework includes two components: the SQL Query Generation Component and the Query Template Identification component. Moreover, the SQL Query Generation Component includes the warm-up phase and the query-generation phase. Thus, in the scalability experiments, we split the running time into three parts: QTI Time, Warm-up Time and Generate Time. In this section, we investigate the impact of each part's running time by varing the number of columns in the relevant table $R$ and the number of rows in the training table $D$. 

 \begin{figure}[t]
  \vspace{-1em}
  \centering
  \includegraphics[width=\linewidth]{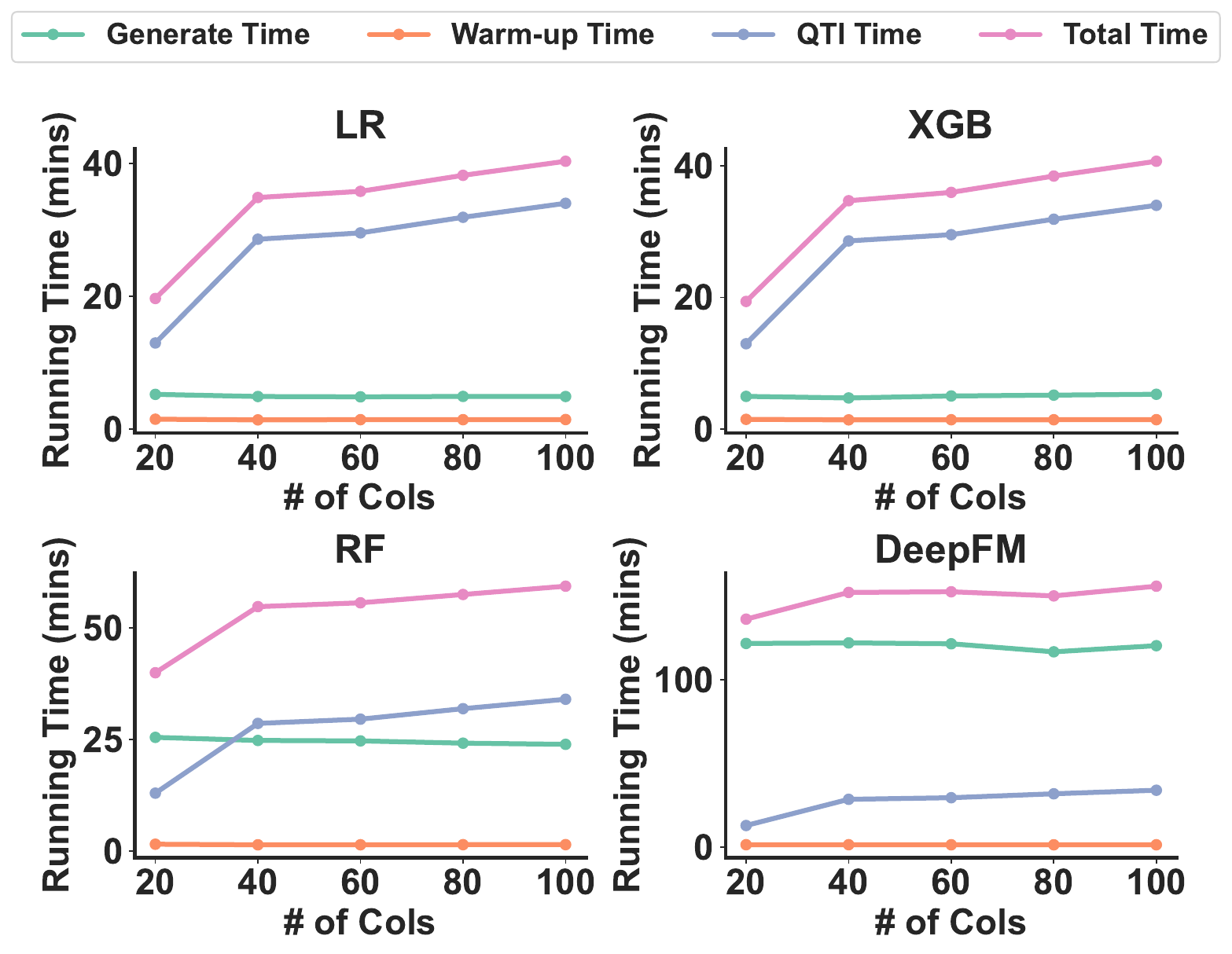}
  \caption{The running time of \sqlgen by varying the number of columns in relevant table $R$ on \textit{Student} dataset. "QTI Time": the running time of the Query Template Identification component. "Warm-up Time": the running time of the warm-up phase in the SQL Query Generation component. "Generate Time": the running time of the query-generation phase in the SQL Query Generation component.}
  \label{fig:col_scalability} 
\end{figure}

\subsubsection{Varying Number of Columns in Relevant Table $R$} 
Since the datasets we utilized do not include wide relevant table with larger than 20 columns, we increase the number of columns by duplicating the original datasets horizontally. For example, we duplicate the \textit{Student} dataset by 13 times to generate a new dataset \textit{Student-Wide} with 130 columns. Because of the space limitation, we present the trend of running time on the \textit{Student-Wide} dataset, which is shown in Figure~\ref{fig:col_scalability}. Results for other datasets can be found in our technical report~\cite{FeatAug_Report}. We have three interesting observations from Figure~\ref{fig:col_scalability}.

Firstly, the Query Template Identification Time does not strictly increase by linear when the number of columns in $R$ increases. Recall that given $attr$, which is a set of attribute from where we select promising attribute combinations (i.e. query templates), the cost of calculating the effectiveness of each query template $T \in \mathcal{S}_{attr}$ is different. That is because the size of $Q_T$ is different for each $T \in \mathcal{S}_{attr}$ and the execution cost of each SQL query $q \in Q_T$ varies. Our cost analysis in Section VI.C indicates \textbf{the maximum cost} of query template identification \textbf{increase linearly} with the increased number of columns in $R$. However in practice, even though the time cost does not increase linearly when the column number in $R$ increases, it is still reasonable.
Secondly, the Warm-up Time and the Generate Time keeps stable no matter how the number of columns changes. The Warm-up Time is mainly determined by the number of iterations used for warming-up when the query template is fixed. The Generate Time includes the model training time, which is highly related to the size of training table rather than the relevant table. Finally, if the downstreaming ML is DeepFM, the Generate Time including model training time becomes the bottleneck on \textit{Student} dataset. That is because compared to other models, DeepFM crosses features automatically, which is time-consuming. 

\textbf{Remark.} Whether the Genarate Time becomes the bottleneck is determined by the dataset and the downstream ML model. With the same training data size, the complexity of the downstream ML model impacts the Generate Time, as Figure~\ref{fig:col_scalability} shows. With the same downstream ML model, the training data size impacts the Genarate Time, as Figure~\ref{fig:row_scalability} shows. Several works utilize data sampling strategy such as coresets~\cite{goodcore, coresetFeatRich, clustercoreset} to reduce the Generate Time. However, these optimization problems and techniques are orthogonal to our problem, which focuses on generate effective predicate-aware SQL queries rather than speeding up the training process of downstream ML models. 

\subsubsection{Varying Number of Rows in Training Table $D$} 
\begin{figure*}[t]
\vspace{0em}
\centering
\subfloat[The running time of \sqlgen by varying the number of rows in training table $D$ on \textit{Tmall}.]{\includegraphics[width=.45\linewidth]{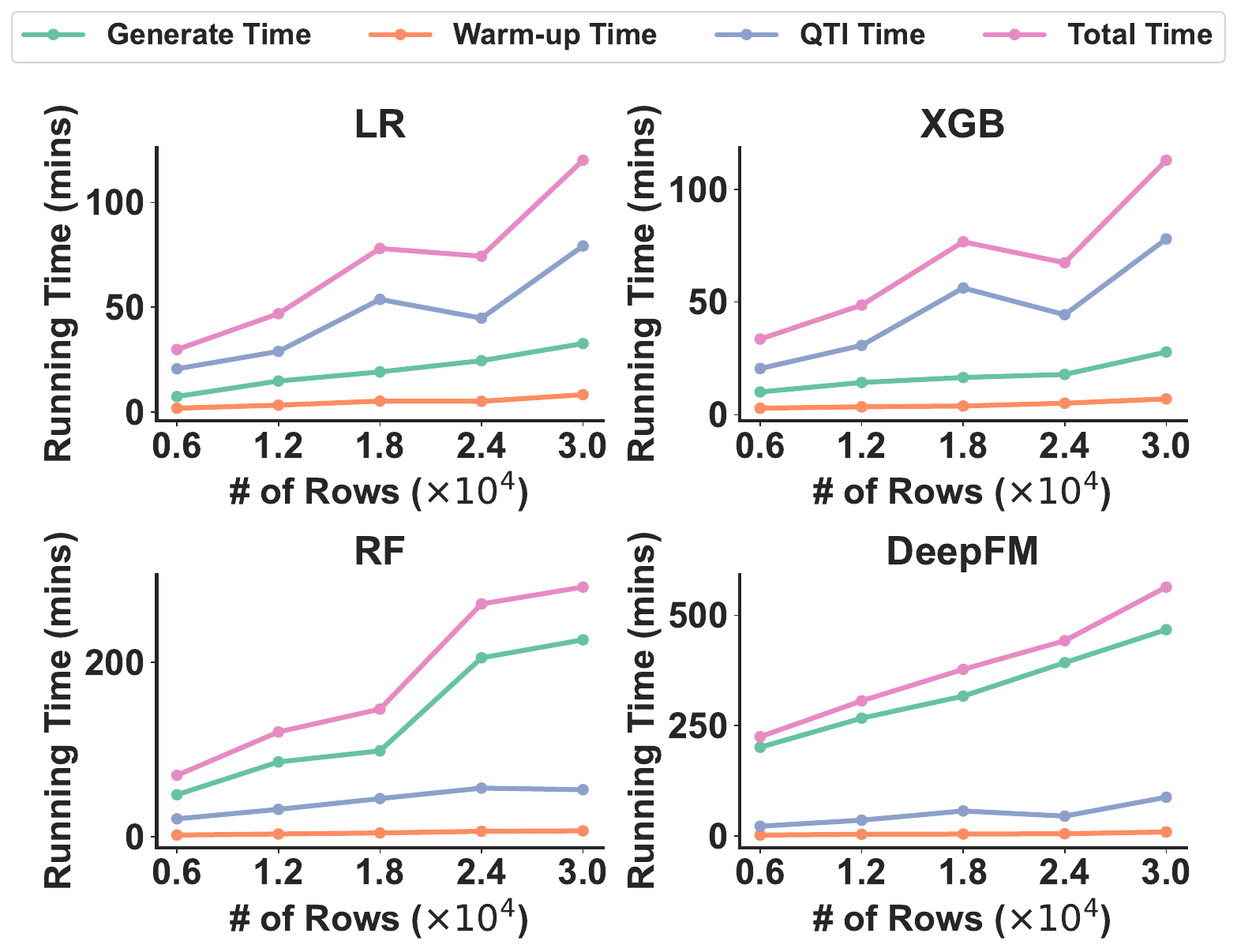} \label{fig:row_scalability_tmall}}\hspace{0pt}
\subfloat[The running time of \sqlgen by varying the number of rows in training table $D$ on \textit{Instarcart}.]{\includegraphics[width=.45\linewidth]{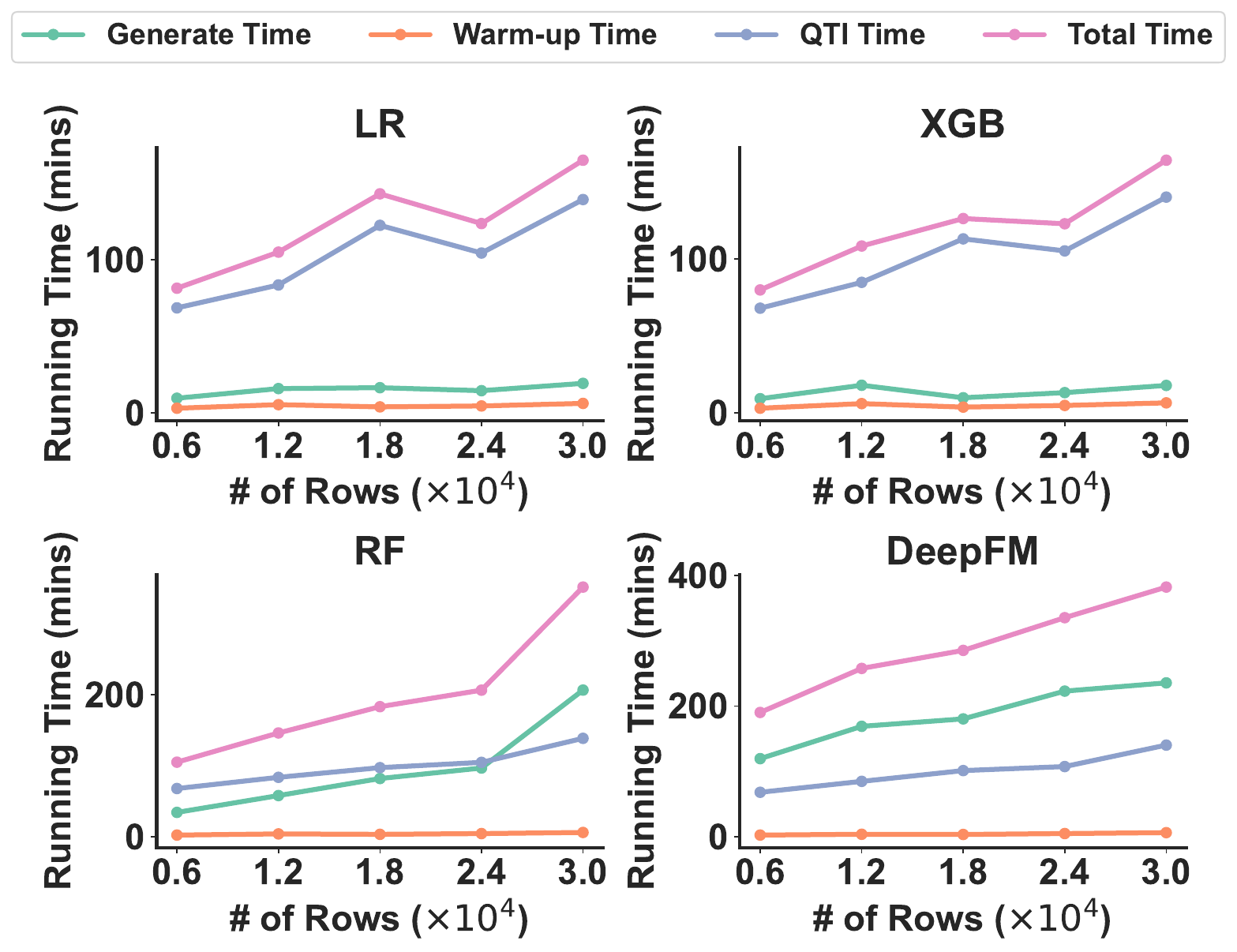}
\label{fig:row_scalability_instacart}}\hspace{0pt}
\subfloat[The running time of \sqlgen by varying the number of rows in training table $D$ on \textit{Student}.]{\includegraphics[width=.45\linewidth]{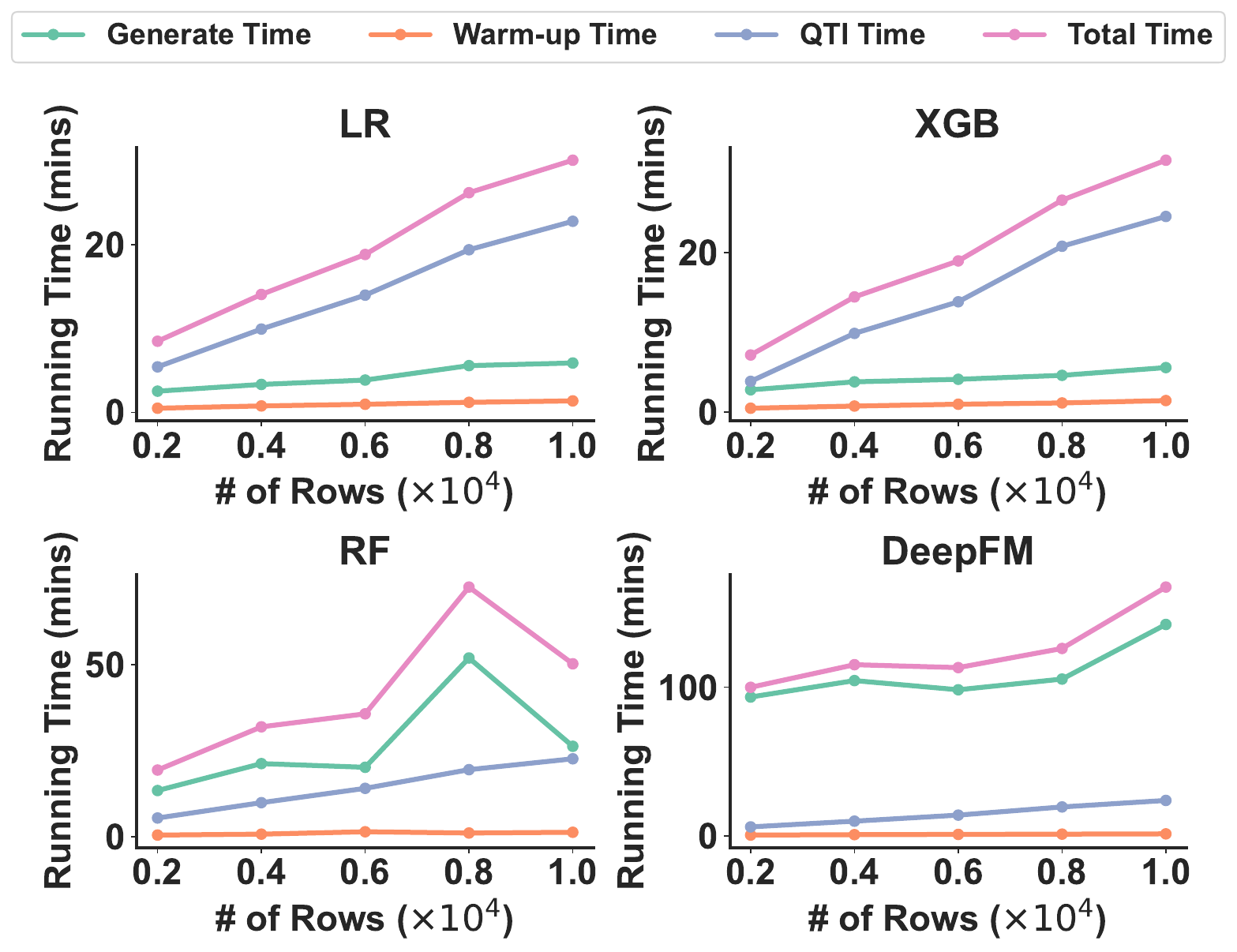}
\label{fig:row_scalability_student}}\hspace{0pt}
\subfloat[The running time of \sqlgen by varying the number of rows in training table $D$ on \textit{Merchant}.]{\includegraphics[width=.45\linewidth]{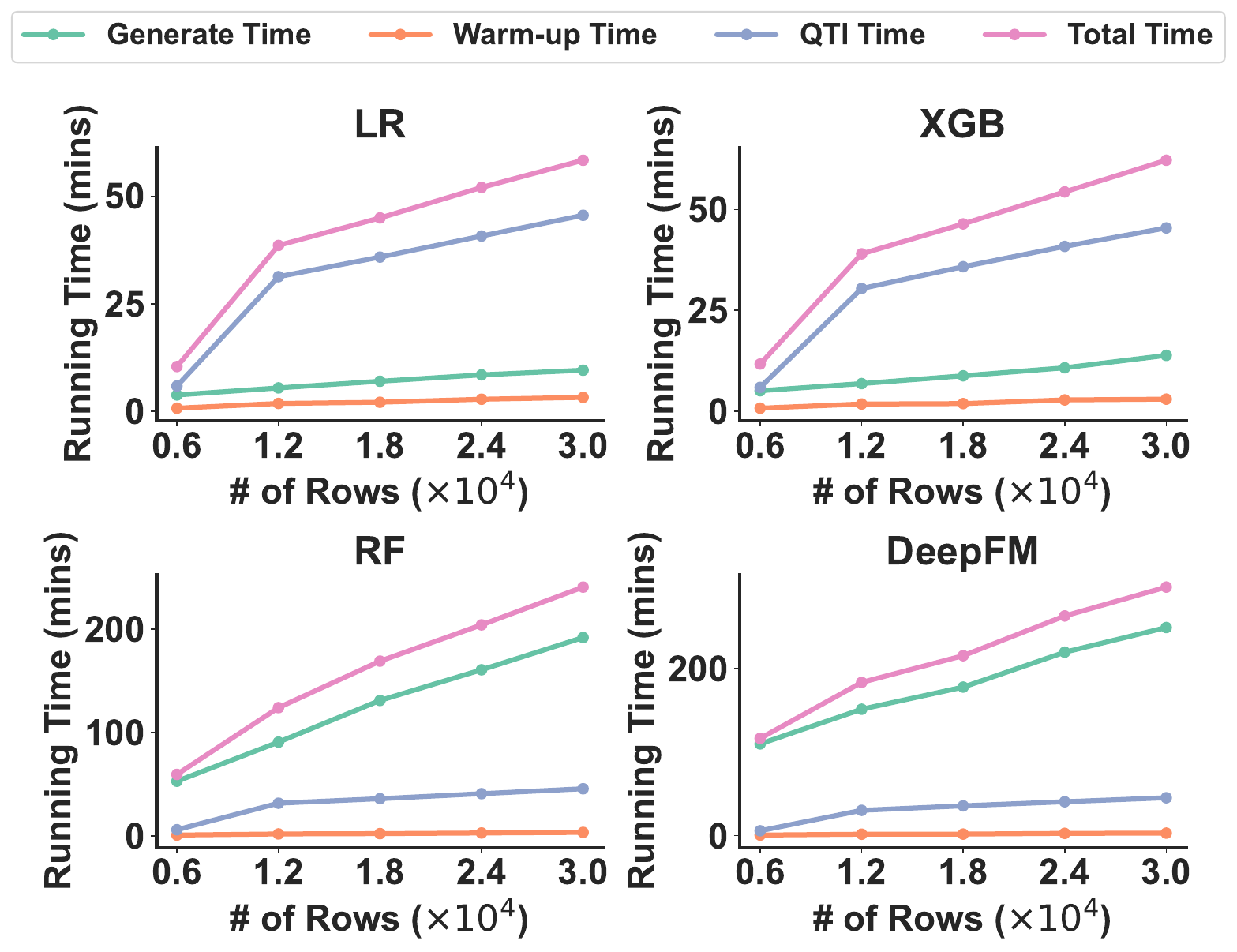}
\label{fig:row_scalability_merchant}}\hspace{0pt}
\caption{The running time of \sqlgen by varying the number of rows in training table $D$. "QTI Time": the running time of the Query Template Identification component. "Warm-up Time": the running time of the warm-up phase in the SQL Query Generation component. "Generate Time": the running time of the query-generation phase in the SQL Query Generation component.}
\label{fig:row_scalability} 
\end{figure*}

Because of the space limitation, we present the trend of running time on \textit{Merchant} dataset by varing the number of rows in training table $D$, which is shown in Figure~\ref{fig:row_scalability}. Results for other datasets can be found in our technical report~\cite{FeatAug_Report}. We have three interesting observations from Figure~\ref{fig:row_scalability}. 


Firstly, the Warm-up time increases linearly when the number of rows in $D$ increases. That is because the Warm-up Phase runs TPE on the related low-cost task such as optimizing MI values and the calculation time of MI values is impacted linearly by the number of rows in training table $D$. Secondly, the Query Template Identification time increases linearly after the number of rows in $D$ is greater than 12k. Thirdly, the increase of the Generate Time is not always linear w.r.t. the increase of the number of rows in $D$. That is because the Generate Time is mainly occupied by the ML model training time, which depends on the complexity of ML model and the training data. The LR model employs the \textit{Ordinary Least Squares (OLS)} has linear time complexity w.r.t. the number of rows in $D$, while the time complexity of other models w.r.t the number of rows in $D$ is not linear.

\subsubsection{\textcolor{blue}{Varying Number of Rows in Relevant Table $R$}}
\begin{figure*}[t]
\vspace{0em}
\centering
\subfloat[\textcolor{blue}{The running time of \sqlgen by varying the number of rows in relevant table $R$ on \textit{Student}.}]{\includegraphics[width=.45\linewidth]{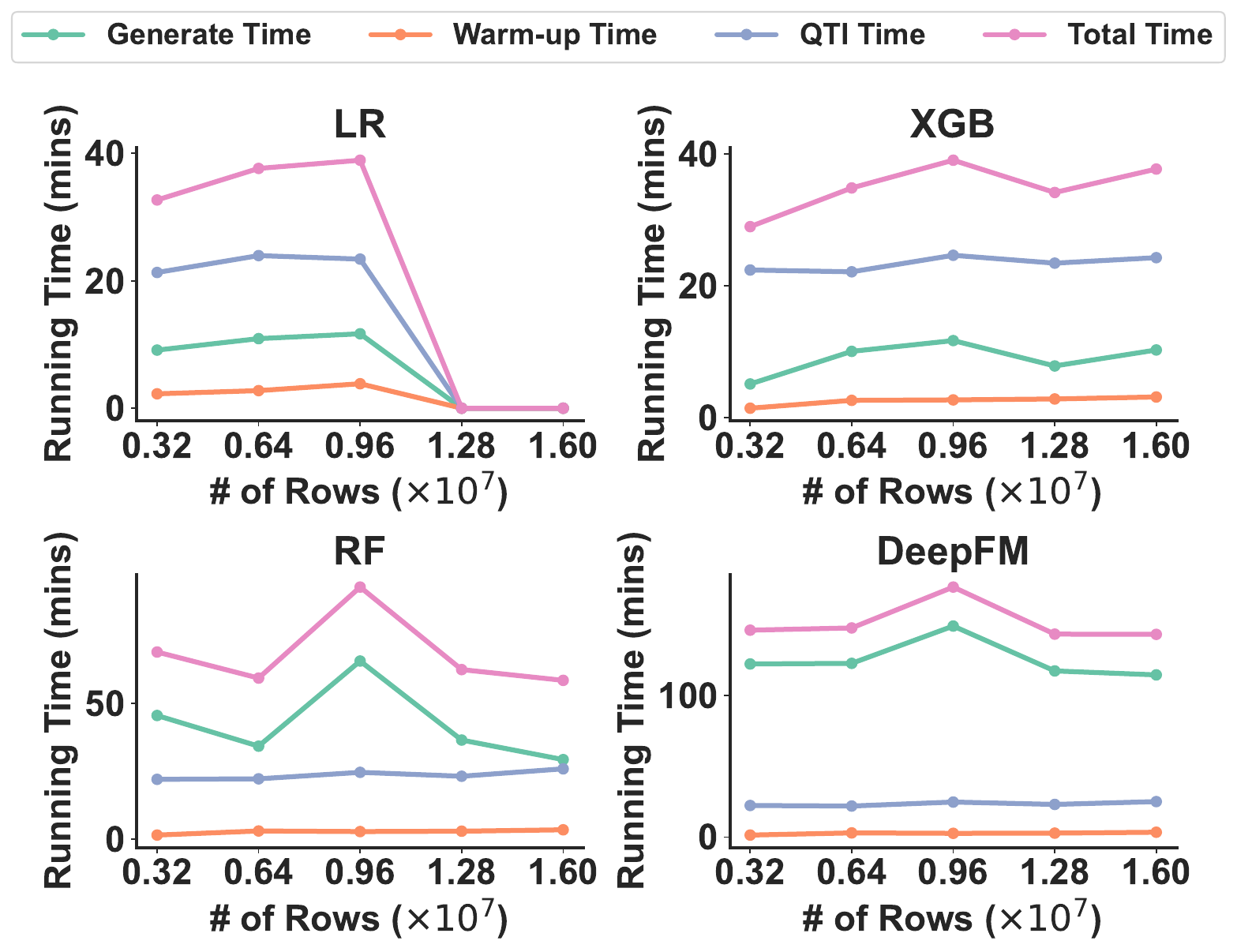} \label{fig:row_scalability_relevant_table_student}}\hspace{0pt}
\subfloat[\textcolor{blue}{The running time of \sqlgen by varying the number of rows in relevant table $R$ on \textit{Merchant}.}]{\includegraphics[width=.45\linewidth]{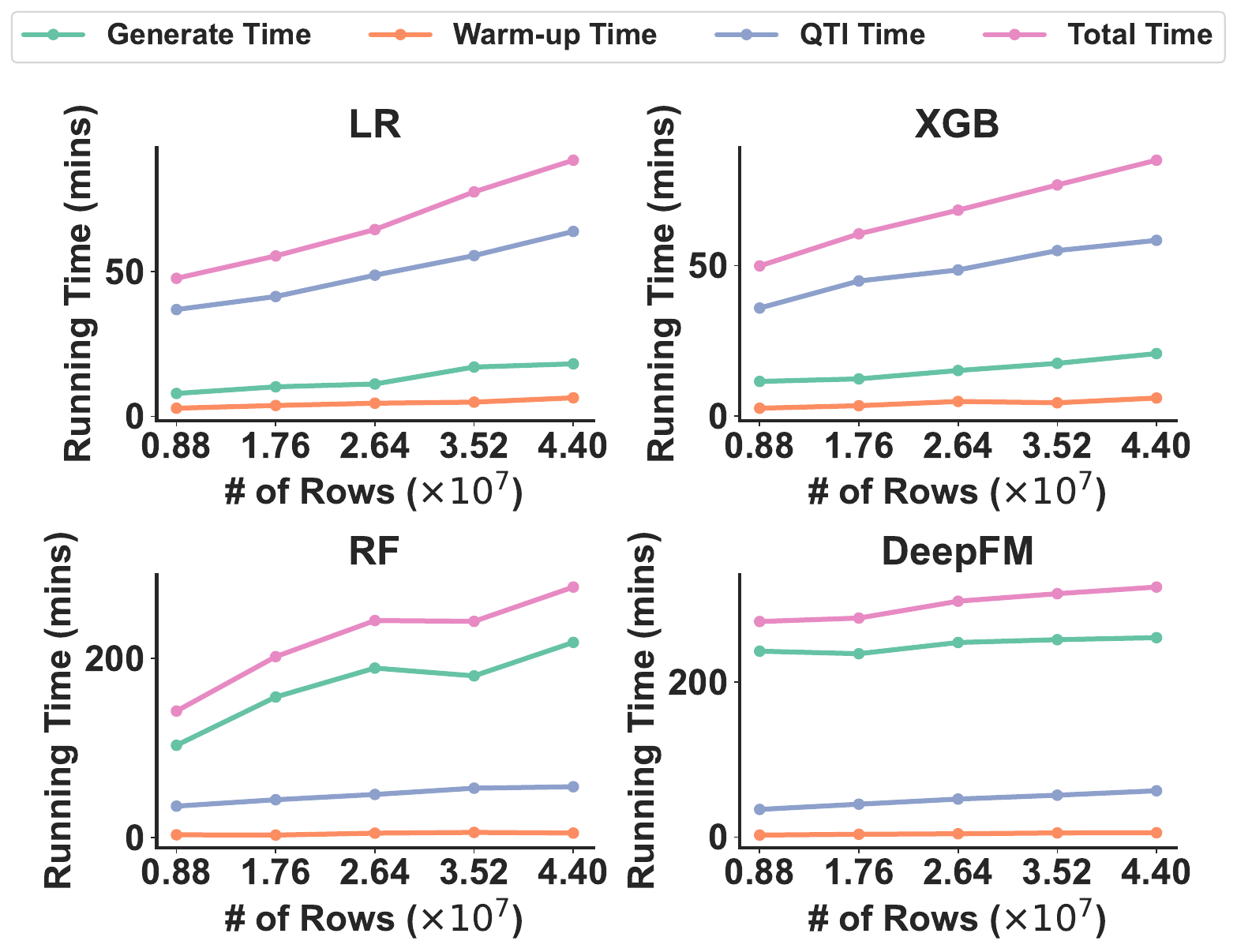}
\label{fig:row_scalability_relevant_table_merchant}}\hspace{0pt}
\caption{\textcolor{blue}{The running time of \sqlgen by varying the number of rows in relevant table $R$. "QTI Time": the running time of the Query Template Identification component. "Warm-up Time": the running time of the warm-up phase in the SQL Query Generation component. "Generate Time": the running time of the query-generation phase in the SQL Query Generation component.}}
\label{fig:row_scalability_relevant_table} 
\end{figure*}

\textcolor{blue}{Because of the space limitation, we present the trend of running time on \textit{Merchant} dataset by varing the number of rows in relevant table $R$, which is shown in Figure~\ref{fig:row_scalability_relevant_table}. Results for other datasets can be found in our technical report~\cite{FeatAug_Report}. We have two interesting observations from Figure~\ref{fig:row_scalability}.}


\textcolor{blue}{Firstly, the Warm-up time and the Query Template Identification time increases linearly when the number of rows in $R$ increases linearly when the number of rows in $R$ increases. That is because the two phases run TPE on the related low-cost task such as optimizing MI values. When the number of rows in training table $D$ keeps stable, the running time of the two phases is impacted by SQL query execution time, which linearly increases when the number of rows in $R$ increases.  Secondly, the increase of the Generate Time is not always linear w.r.t. the increase of the number of rows in $R$. That is because the Generate Time is mainly occupied by the ML model training time, which depends on the complexity of ML model and the training data.}



\vspace{-.3em}

\section{Conclusion}
In this paper, we have addressed the challenging problem of augmenting features automatically from one-to-many relationship tables, a task critical for enhancing the performance of machine learning models. Our proposed framework, \sqlgen leverages the power of effective predicate-aware SQL queries to enrich feature sets. We discussed how to extend a widely used Hyperparameter Optimization (HPO) algorithm TPE to our problem and enhance it by warming up the search process.
Furthermore, to make \sqlgen more practical, we discuss how to identify promising query templates. The beam search idea can partially fit for this problem with exponential complexity. By incorporating the low-cost proxy for query template effectiveness and prediction of promising query templates with a ML model, the cost of initial beam search is highly reduced.
We conducted extensive experiments using four real-world ML datasets to evaluate \sqlgen and compare it with the popular \textit{Featuretools}. The results shows that \sqlgen consistently outperformed the established \textit{Featuretools} framework, which means that show that \sqlgen was able to discover more effective features by constructing effective predicate-aware SQL queries. Our future work will venture into more complex scenarios involving mixed relationships in database tables, such as one-to-one and many-to-many associations. This expansion will cover a broader range of real-world applications. 

\bibliographystyle{IEEEtran}
\bibliography{reference}

\end{document}